\documentclass{article} 
\usepackage[preprint]{colm2026_conference}

\usepackage{microtype}
\usepackage{hyperref}
\usepackage{url}
\usepackage{booktabs}
\usepackage[table]{xcolor}
\usepackage{subcaption}
\usepackage{placeins}
\usepackage{tabularx}
\usepackage{booktabs}
\usepackage{ragged2e}
\PassOptionsToPackage{numbers,sort&compress}{natbib}

\usepackage{silence}
\definecolor{cornflowerblue}{RGB}{100, 149, 237}

\definecolor{amethyst}{rgb}{0.6, 0.4, 0.8}
\definecolor{tealgreen}{RGB}{0, 153, 120}

\usepackage{lineno}

\definecolor{darkblue}{rgb}{0, 0, 0.5}
\hypersetup{colorlinks=true, citecolor=darkblue, linkcolor=darkblue, urlcolor=darkblue}

\usepackage[ampersand]{easylist}
\ListProperties(Hide2=1,Hide3=2,Progressive*=.5cm,Numbers3=l, Numbers4=r,FinalMark3={)})
\usepackage{etoolbox}
\AtBeginEnvironment{easylist}{\ListProperties(Start1=1)}

\usepackage{xcolor}

\definecolor{cblue}{RGB}{8, 85, 153}

\usepackage{amsmath}
\usepackage[capitalise]{cleveref}
\crefname{equation}{Eq.}{Eqs.}
\Crefname{equation}{Eq.}{Eqs.} 
\crefname{figure}{Fig.}{Figs.}
\Crefname{figure}{Fig.}{Figs.}
\crefname{section}{Sec.}{Secs.}
\Crefname{section}{Sec.}{Secs.}
\usepackage{float}
\usepackage{placeins} 
\usepackage{adjustbox}

\usepackage{makecell}
\usepackage[ampersand]{easylist}
\usepackage{todonotes}
\usepackage{caption}
\usepackage{xspace}
\usepackage{markdown}
\usepackage{multirow}
\usepackage{multicol}

\usepackage{adjustbox}

\newcommand{\siglow}{Passed both checks}
\newcommand{\nonsiglow}{Failed the Yes check}
\newcommand{\sighigh}{Failed the Overlap check}
\newcommand{\nonsighigh}{Failed both checks}

\usepackage{pifont}

\newcommand{\cmark}{\textcolor{green}{\ding{51}}}
\newcommand{\xmark}{\textcolor{red}{\ding{55}}}

\title{Sanity Checks for Agentic Data Science}

\author{Zachary T. Rewolinski\thanks{Equal contribution}, \:Austin V. Zane\footnotemark[1]\\
Dept. of Statistics\\
UC Berkeley\\
\AND
Hao Huang\\
Dept. of EECS\\
UC Berkeley\\
\AND
Chandan Singh, Chenglong Wang, Jianfeng Gao \\
Microsoft Research\\
\AND
Bin Yu \\
Dept. of Statistics, Dept. of EECS, Center for Computational Biology\\
UC Berkeley\\
}

\begin{document}

\ifcolmsubmission
\linenumbers
\fi

\maketitle

\begin{abstract}
Agentic data science (ADS) pipelines have grown rapidly in both capability and adoption, with systems such as OpenAI Codex now able to directly analyze datasets and produce answers to statistical questions.
However, these systems can reach falsely optimistic conclusions that are difficult for users to detect. To address this, we propose a pair of lightweight sanity checks grounded in the Predictability-Computability-Stability (PCS) framework for veridical data science. 
These checks use reasonable perturbations to screen whether an agent can reliably distinguish signal from noise, acting as a falsifiability constraint that can expose affirmative conclusions as unsupported.
Together, the two checks characterize the trustworthiness of an ADS output, e.g. whether it has found stable signal, is responding to noise, or is sensitive to incidental aspects of the input.
We validate the approach on synthetic data with controlled signal-to-noise ratios, confirming that the sanity checks track ground-truth signal strength.
We then demonstrate the checks on 11 real-world datasets using OpenAI Codex, characterizing the trustworthiness of each conclusion and finding that in 6 of the datasets an affirmative conclusion is not well-supported, even though a single ADS run may support one. 
We further analyze failure modes of ADS systems and find that ADS self-reported confidence is poorly calibrated to the empirical stability of its conclusions\footnote{All relevant code can be found at https://github.com/zachrewolinski/stat-genie}.
\end{abstract}

\section{Introduction}
\label{sec:intro}

Agentic data-science (ADS) pipelines are becoming increasingly powerful~\citep{guo2024ds,nam2025ds}, driven in part by the rapid spread of autonomous systems such as OpenAI Codex and Claude Code.
These pipelines are capable of taking as input a natural-language query and a relevant dataset and executing the full analytical workflow to produce an answer, enabling end-to-end ``vibe data science.''
Any user can obtain a polished analysis, complete with model selection, hypothesis tests, and a written summary.
However, the conclusions produced by ADS pipelines can be wrong in ways that are difficult to detect.
%
%
An agent may make an unreasonable data-cleaning decision,
choose an inappropriate model,
or overlook confounders, but the resulting report will still appear polished and defensible~\citep{bertran2026many}.
Worse, for difficult datasets, a handful of runs can yield an entirely different conclusion than a full stability analysis, underscoring the need for systematic checks (\cref{fig:kde_main}).
Thus, the rapid adoption of ADS pipelines poses an urgent risk to flooding science with unreliable findings~\citep{tan2026mirage}.
%

%
%
%
Existing solutions to tackle this problem have limitations.
Classical tests of statistical significance generally fail in agentic settings. For example, we observe that agents handle statistical evidence poorly in practice, over-interpreting estimates that lack significance and ``$p$-hacking'' by anchoring on whichever test is most favorable (\cref{subsec:null-strong-yes}).
Alternatively, simply asking an agent to assess the uncertainty of an analysis also does not succeed, as agent-reported confidence is poorly calibrated to the actual stability of an underlying result (\cref{subsec:agent-uncertainty-estimation}).
Finally, ``multiverse'' analyses enable quantifying the impact of different choices in an ADS pipeline, but their outputs quickly become too large for human review, creating a need for actionable checks~\citep{bertran2026many,rohrer2026multiverse}.

To tackle this challenge, we make use of the Predictability-Computability-Stability (PCS) framework for veridical data science~\citep{yu2020veridical,yu2024veridical,rewolinski2025pcs}.
It establishes that trustworthy, reality-checked conclusions must be stable under reasonable perturbations across different data science stages, such as problem specification, data cleaning, and model choices.
In this spirit, we generate small, reasonable changes, which we call \textit{PCS perturbations}.
These perturbations probe agentic failure modes which would not impact a competent human analyst's conclusions.
If an agent's conclusions change dramatically under these perturbations, they are not trustworthy.

\begin{figure}[t]
    \centering
    \includegraphics[width=\linewidth]{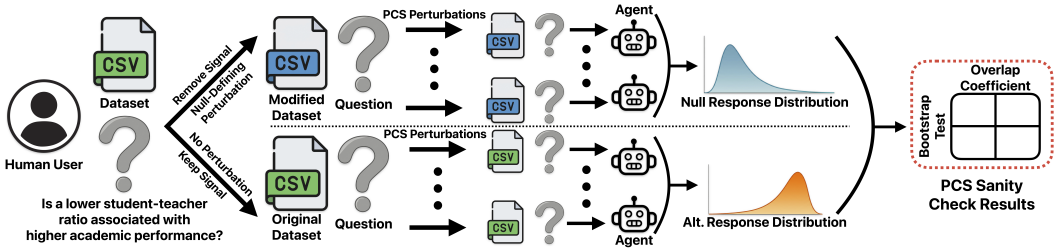}
    \caption{Pipeline for computing the PCS sanity checks. The user supplies a dataset and question. The pipeline then generates perturbed variants of these under both null-defining and PCS perturbations. The coding agent analyzes each independently, producing null and alternative response~distributions. A bootstrap test \& overlap coefficient then inform whether the agent found a data-grounded, stable positive conclusion.}
    \label{fig:schematic}
\end{figure}

Consistency under these perturbations is necessary but not sufficient, since an agent could also arrive at a stable but biased conclusion. 
Since ground-truth conclusions are rarely known in data science, we use a \textit{null-defining perturbation} to put the agent in a setting where the answer is known and testable.
For example, we may remove all the signal in the data (e.g., independently shuffling each column) to establish a negative control against which we can measure the agent's capabilities.
We then run the agent many times to generate two empirical distributions: one where both null-defining and PCS perturbations are applied, which we call the \textit{null distribution}, and another where only PCS perturbations are applied, which we call the \textit{alternative distribution}.
%
%
Using these two distributions, we apply two complementary sanity checks (see \cref{fig:schematic}):

(1) \textit{The Yes check}: a one-sample bootstrap test of whether the ADS mean response on the original data is significantly above the midpoint of the scale (i.e., in the ``Yes" range), thereby testing whether a positive conclusion was reached by chance.

(2) \textit{The Overlap check}: quantifies the similarity between the null and alternative distributions, thereby testing for bias in the ADS conclusion that does not come from signal in the data.

%
%
%
%
In experiments using OpenAI Codex,
%
we first validate the approach on simulations with controlled signal-to-noise ratios, confirming that the checks track ground-truth signal strength (\cref{subsec:pve-results}).
%
We then apply them to eleven real-world datasets from the BLADE benchmark, where the checks characterize each dataset's signal regime and reveal one case in which the agent's affirmative conclusion is not well-separated from its behavior on pure noise (\cref{subsec:main-results}). 
We further analyze failure modes of ADS systems (\cref{sec:analysis}), showing that ADS self-reported confidence is poorly calibrated to the empirical stability of its conclusions and characterizing how different steps fail using the BLADE agentic harness (an alternative to Codex).
%
%
%
%
%
%
We view this work as a first step toward evaluation standards for agentic data science, grounded in the core principles of the PCS framework for veridical data science.

\section{Related Work}

\textbf{Agentic data science (ADS).}
The risks of AI automation in the scientific workflows have been documented in fields adjacent to data science, including LLM-hacking for annotation~\citep{baumann2025large,tseng2025evaluating} and broader stages of the research pipeline~\citep{luo2025more}.
More directly relevant, \citet{asher2026sycophancy} demonstrates that guardrails are necessary to prevent humans from $p$-hacking using LLM-driven data science, while other work highlights the value of transparency in intermediate data-science artifacts as a partial remedy~\citep{sivaraman2026meansendsupportingreasoning,nam2025ds}.
Despite these warning signs, much of the ADS literature has focused on capability and benchmarking, including evaluation suites such as DSGym~\citep{nie2026dsgym}, ScienceAgentBench~\citep{chen2024scienceagentbench}, IDA-Bench~\citep{li2025ida} and others~\citep{song2025evaluating,li2025autosdt}, alongside accompanying agentic frameworks~\citep{guo2024ds}.
Comparatively, little attention has been paid to whether ADS conclusions can be trusted, motivating the need for intuitive and actionable sanity checks.

\textbf{Perturbation-based analysis.}
The PCS framework~\citep{yu2020veridical,yu2024veridical} and its use of generative AI~\citep{rewolinski2025pcs} has been shown to work well in solving scientific problems~\citep{basu2018Iterative,abbasi2018deeptune,wang2024epistasis,tang2025simplified,wu2016stability,antonello2024generative} and generalizes various perturbations or viewpoints common in the literature, e.g. the multiverse analysis~\citep{steegen2016increasing,del2021traveler},
sensitivity analysis~\citep{christopher2002identification},
and others~\citep{fisher2019all}, along with accompanying software~\citep{liu2020boba,sarma2024milliways,duncan2022veridicalflow}.
Recent work has proposed the promising use of multiverse analyses for reporting ADS conclusions~\citep{bertran2026many}, but the resulting multiverse is often too large for human review, and thus its nature is more descriptive than actionable.
This creates a need for actionable sanity checks which evaluate conclusion \textit{stability}.

\textbf{Uncertainty quantification for language models.}
Many frameworks for uncertainty quantification in traditional statistical learning have been configured for language model applications. This includes conformal inference~\citep{quach2024conformal,noorani2025conformal}, confidence intervals~\citep{wu2026efficientevaluationllmperformance}, and Bayesian approaches~\citep{ross2025textualbayesquantifyinguncertainty}. It also includes more holistic approaches, such as automatic evaluation of an LLM-generated plan\citep{seo2026assumptionsactionsturningllm} and participatory refinement~\citep{delacroix2025beyond}.
However, these approaches have not been tailored to data science, which contains a compounding variety of judgement calls made by the user~\citep{gelman2013garden,yu2020veridical}.

\section{Methods}
\label{sec:method}

We consider the setting where a user supplies an agentic system with a dataset and a yes or no question. 
This setup can produce qualitative insights for a wide range of questions.
%


\subsection{Generating response distributions using perturbations}
\label{subsec:perturbations}

As discussed in \cref{sec:intro}, PCS posits that trustworthy conclusions must be stable under reasonable perturbations to the data and the problem formulation.
We~define two perturbation types: one that modifies the data's signal, and one that leaves it unchanged.
Combined, they provide a useful sanity check on the trustworthiness of an agent's conclusion.

\textbf{Null-defining perturbations.}
Evaluating ADS conclusions is difficult without ground truth.
We address this by examining responses in settings where the answer is known by removing signal from the data, similar to negative control methods~\citep{shi2020selective}.
By breaking the association between features and outcomes, any inferred relationship should disappear, creating a distribution of conclusions under known null conditions.
We call such modifications ``null-defining perturbations".
Examples include shuffling feature values independently or replacing outcomes with random draws.
Running the agent on the perturbed data yields the empirical \emph{null distribution}.
If no null-defining perturbation is applied, we call the resulting distribution the \emph{alternative distribution}. 

\textbf{Probing failure modes with PCS perturbations.}
To understand the response variation, we probe ADS failure modes using PCS-inspired perturbations to metadata or prompts.
These ``PCS perturbations" (\cref{tab:pcs-perturbations}) leave data signal unchanged and should not affect experienced data scientists.
We apply these PCS perturbations when computing both for the null and alternative distributions to assess stability.

\newcolumntype{Y}{>{\RaggedRight\arraybackslash}X}

\begin{table}[t]
\centering
\caption{Perturbations used in the PCS sanity checks.
The null-defining distribution breaks associations between the features and the outcome.
The PCS perturbations should not affect an experienced data scientist or agentic data science system, although some perturbations (e.g. \textit{positive leading statement}) may mislead human data scientists due to confirmation bias.
}
\small
\renewcommand{\arraystretch}{1.5} 
\begin{tabularx}{\textwidth}{@{} >{\hsize=0.6\hsize}Y >{\hsize=0.3\hsize}Y >{\hsize=1.3\hsize}Y >{\hsize=0.8\hsize}Y @{}}
\toprule
\textbf{Perturbation} & \textbf{Type} & \textbf{Description} & \textbf{Failure Mode Tested} \\
\midrule
Shuffle Feature Values & Null-Defining & Shuffles each column independently to break associations in the data. & Incorrect data analysis conclusions. \\ \addlinespace
Add Non-Signal Features & PCS & Appends features irrelevant to \& uncorrelated with the task of interest. & Distraction by irrelevant covariates. \\ \addlinespace
Anonymize Feature Names & PCS & Replaces all column names with generic identifiers (e.g., \texttt{feature1}, \texttt{feature2}). & Reliance on semantic variable names from pre-training data. \\ \addlinespace
Shuffle Feature Names & PCS & Applies a random permutation of existing column names while leaving data values and feature descriptions in metadata unchanged. & Robustness to misleading variable names. \\ \addlinespace
Positive Leading Statement & PCS & Prepends the question with a strong prior belief that the answer is ``Yes.'' & Sycophancy toward user-stated beliefs. \\ \addlinespace
Negative Leading Statement & PCS & Prepends the question with a strong prior belief that the answer is ``No.'' & Sycophancy toward user-stated beliefs. \\
\bottomrule
\end{tabularx}
\label{tab:pcs-perturbations}
\end{table}

\textbf{Eliciting continuous responses.}
We prompt the agent to respond on a scale from 0 (strong ``No'') to 100 (strong ``Yes'').
While binary outputs still permit stability analysis, restricting the agent's output space discards information about the strength of its conviction.

\subsection{Two PCS sanity checks}
\label{subsec:sanity-checks}
Given the null and alternative distributions, we want to answer two questions.
First, does the agent consistently affirm the research question on the real data, or could its positive response be a fluke?
Second, does the agent behave differently on real data than on noise, or does it produce similar outputs regardless of whether signal is present?
Each question targets a failure mode that the other misses.

\textbf{\textit{The Yes Check}. Is the agent consistently affirmative?}
A single high score could be a fluke.
We therefore test whether the agent's mean response on the original data consistently exceeds the scale midpoint (50) across repeated runs and perturbations.
A mean above 50 indicates net agreement with the research question, while a mean at or below 50 suggests a lack of evidence for a ``Yes'' answer.
Importantly, by aggregating over many runs, the test guards against drawing conclusions from a single lucky (or unlucky) pipeline execution.
To formalize this, we test the one-sided hypothesis $H_0\!:\mu_{\text{alt}}=50$ against $H_1\!:\mu_{\text{alt}}>50$, where $\mu_{\text{alt}}$ is defined as the population mean, drawn from the alternative distribution.
We draw $B$ bootstrap samples from the observed alternative responses, computing the $p$-value as the proportion of bootstrap means that fall at or below 50, applying the \citet{phipson2010permutation} correction to avoid $p$-values of exactly zero.
We also report bootstrap percentile confidence intervals, providing a range of plausible average responses.
This check treats responses across PCS perturbation types as identically distributed. We verify this in Appendix~\ref{appx:check_assump}.

\textbf{\textit{The Overlap check}. Can the agent distinguish signal from noise?}
An agent's average response could be well above 50, yet still produce a response distribution that overlaps heavily with the null, suggesting it cannot distinguish signal from noise.
%
To capture this, we compute the overlap coefficient (OVL) between the alternative \& null densities ($f_{alt}$ \& $f_{null}$): 
\begin{equation} 
    \text{OVL} = \int \min\!\big(f_{\text{alt}}(x),\; f_{\text{null}}(x)\big)\, dx.
\end{equation} 
We estimate densities via Gaussian kernel density estimation with Scott's rule~\citep{scott1992multivariate} for bandwidth selection (the default in most standard scientific computing libraries), and the integral is evaluated numerically over $[0, 100]$.
An OVL near~0 indicates well-separated distributions, while an OVL near~1 indicates that the agent cannot tell real data from noise.

\textbf{Interpreting the PCS sanity checks.}
\label{subsec:interpret-method}
Combining both checks produces a $2 \times 2$ classification of each dataset, summarized in \cref{tab:qmatrix}.
We can binarize each sanity check using a threshold, e.g. for the \textit{Yes check} we assess whether the bootstrap $p$-value is below significance level~$\alpha$, and for the \textit{Overlap check} whether $\text{OVL} < \tau$ for a user-chosen threshold~$\tau$.
%
%
%
Note from \cref{tab:qmatrix} that we obtain different information when one check fails but the other succeeds.
%


\definecolor{lightgrey}{gray}{0.9}

\begin{table}[t]
\centering
\small
\caption{Interpretations of the four PCS sanity-check regimes.}
\renewcommand{\arraystretch}{2.0} 
\begin{tabularx}{\textwidth}{>{\RaggedRight\bfseries}p{2.75cm}|p{4.75cm}Y}
\toprule
\textnormal{} & \centering\textbf{\textit{Overlap check} passed} \cmark & \centering\arraybackslash\textbf{\textit{Overlap check} failed} \xmark \\ \midrule
\textbf{\textit{Yes check} passed} \cmark & The positive conclusion is stable. & The positive conclusion may not be grounded in the data. \\ \midrule
\textbf{\textit{Yes check} failed} \xmark & There is some positive signal, but not enough for a ``Yes". & No evidence supporting a positive conclusion. \\ \bottomrule
\end{tabularx}
\label{tab:qmatrix}
\end{table}
\FloatBarrier
\section{Results}
\label{sec:results}

In this section, we demonstrate that the PCS sanity checks effectively measure simulated signal in the data (\cref{subsec:pve-results}) and report results for each of the BLADE datasets (\cref{subsec:main-results}).

\subsection{Experimental Setup}
\label{subsec:experimental-setup}

\textbf{Datasets.}
We evaluate our approach on eleven datasets from the BLADE benchmark~\citep{gu2024blade}, a collection of tabular datasets paired with binary yes-no questions.
Further details regarding the datasets and their corresponding tasks are provided in Appendix \ref{appx:datasets}.

\textbf{Perturbations.}
To define the null distribution, we break all data associations by shuffling feature values independently.
We apply five PCS perturbations (Table \ref{tab:pcs-perturbations}) 20 times each, creating 100 observations for both the null and alternative distributions per dataset.

\textbf{Agent \& prompt.}
We use GPT‑5.2‑Codex as our agent, set to ``high" reasoning effort.
Prompted as ``an expert data scientist", the agent provides: (1) an integer Likert score, where 0 indicates strong “No” and 100 indicates strong “Yes”, and (2) a text explanation.
The exact prompt is in Appendix \ref{appx:prompt-info}.
We also use the BLADE agentic system~\citep{gu2024blade} for a step-level analysis of where instability enters (\cref{subsec:heatmaps}).

\textbf{Sanity check details.}
The bootstrap test (level $\alpha=0.05$) evaluates if the alternative mean exceeds 50 (i.e., is in the ``Yes" range) using $B = 10{,}000$ resamples.
The overlap coefficient threshold separating low from high distributional overlap is set to $\tau = 0.2$.

\subsection{PCS Sanity Checks Correlate with Signal Strength in Simulations}
\label{subsec:pve-results}

Lacking ground-truth for real-world data, we use simulations to determine the signal~needed for an agent to distinguish it from noise, using the signal-to-noise ratio as a tuning knob.

\textbf{Setup.}
For each BLADE dataset, we identify the dependent variable $Y$ and (one-hot encoded) independent variables $X$, which includes an intercept.
We regress $Y$ on $X$ to fit an OLS model $\widehat{Y} = X\beta$.
We then construct a synthetic dependent variable $Z=\widehat{Y}+\varepsilon$, where the noise variance is calibrated so that the signal explains a target fraction of the total variance.
Specifically, we set 
    $\sigma_{\varepsilon} = \sqrt{\text{Var}(\widehat{Y}) \cdot \frac{1 - \text{PVE}}{\text{PVE}}}$,
%
where $\text{PVE}=\text{Var}\left(\mathbb{E}[Y\mid X]\right)/\text{Var}(Y)$ represents the proportion of variance explained.
At the boundary, $\text{PVE} = 0$ represents pure noise ($Z\sim N(\Bar{Y}, \sigma_Y$)), while $\text{PVE} = 1$ represents pure signal ($Z=\widehat{Y}$).
%
Responses on data with $\text{PVE}\in\{0.0, 0.01, 0.1\}$ serve as the alternative group. The null group is as characterized in \cref{subsec:experimental-setup}.
The PCS perturbations from Table \ref{tab:pcs-perturbations} are applied five times each, creating an alternative distribution with 25 observations.

\begin{figure}[tb]
     \centering
     \begin{subfigure}[b]{0.48\textwidth}
         \centering
         \includegraphics[width=.75\textwidth]{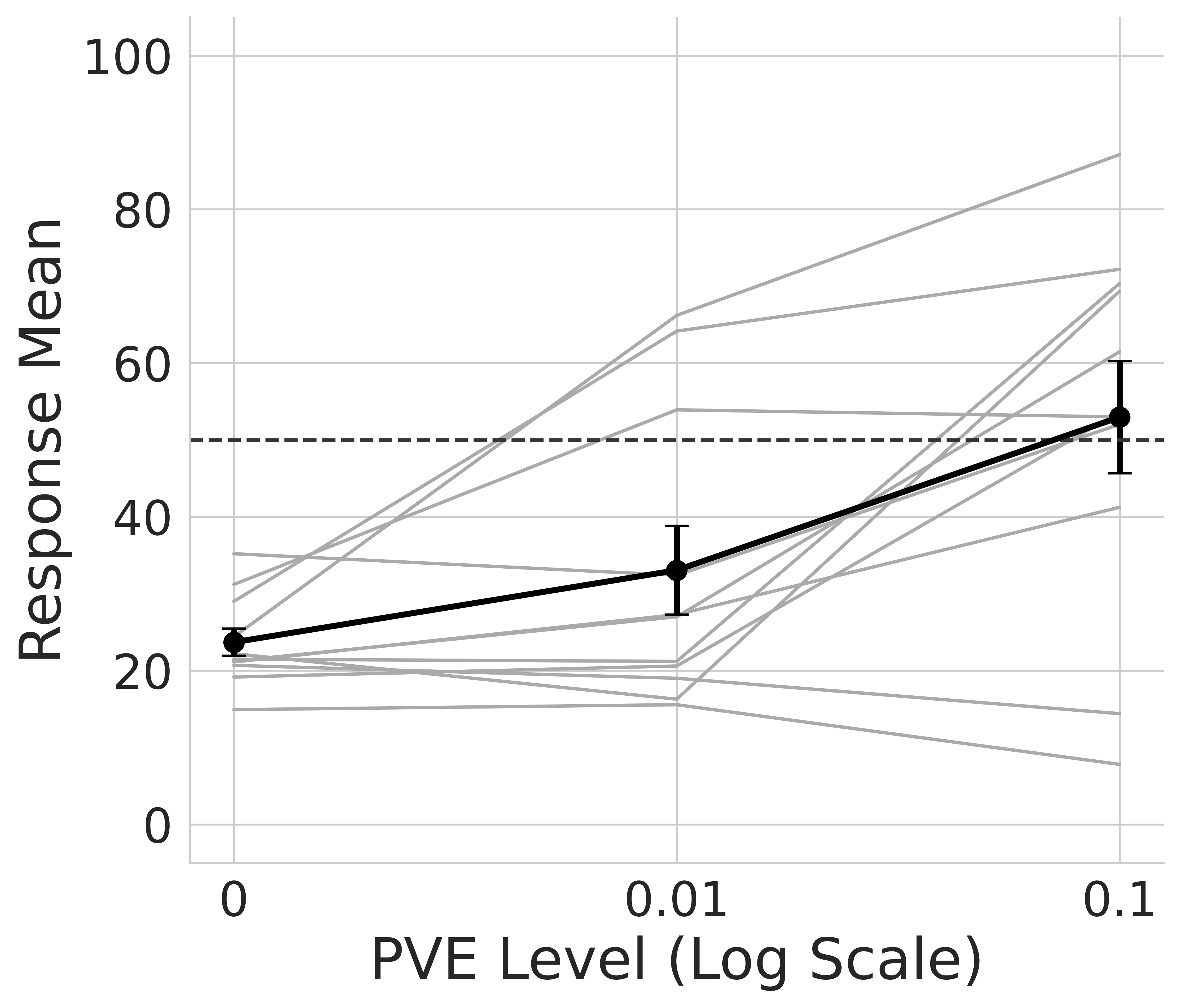}
         \caption{The agent's response increases as we add signal to the data. The two datasets that decrease, \texttt{affairs} and \texttt{reading}, ask directional questions, with the signal going in the opposite direction.}
         \label{fig:pve-trend-response}
     \end{subfigure}
     \hfill 
     \begin{subfigure}[b]{0.48\textwidth}
         \centering
         \includegraphics[width=.75\textwidth]{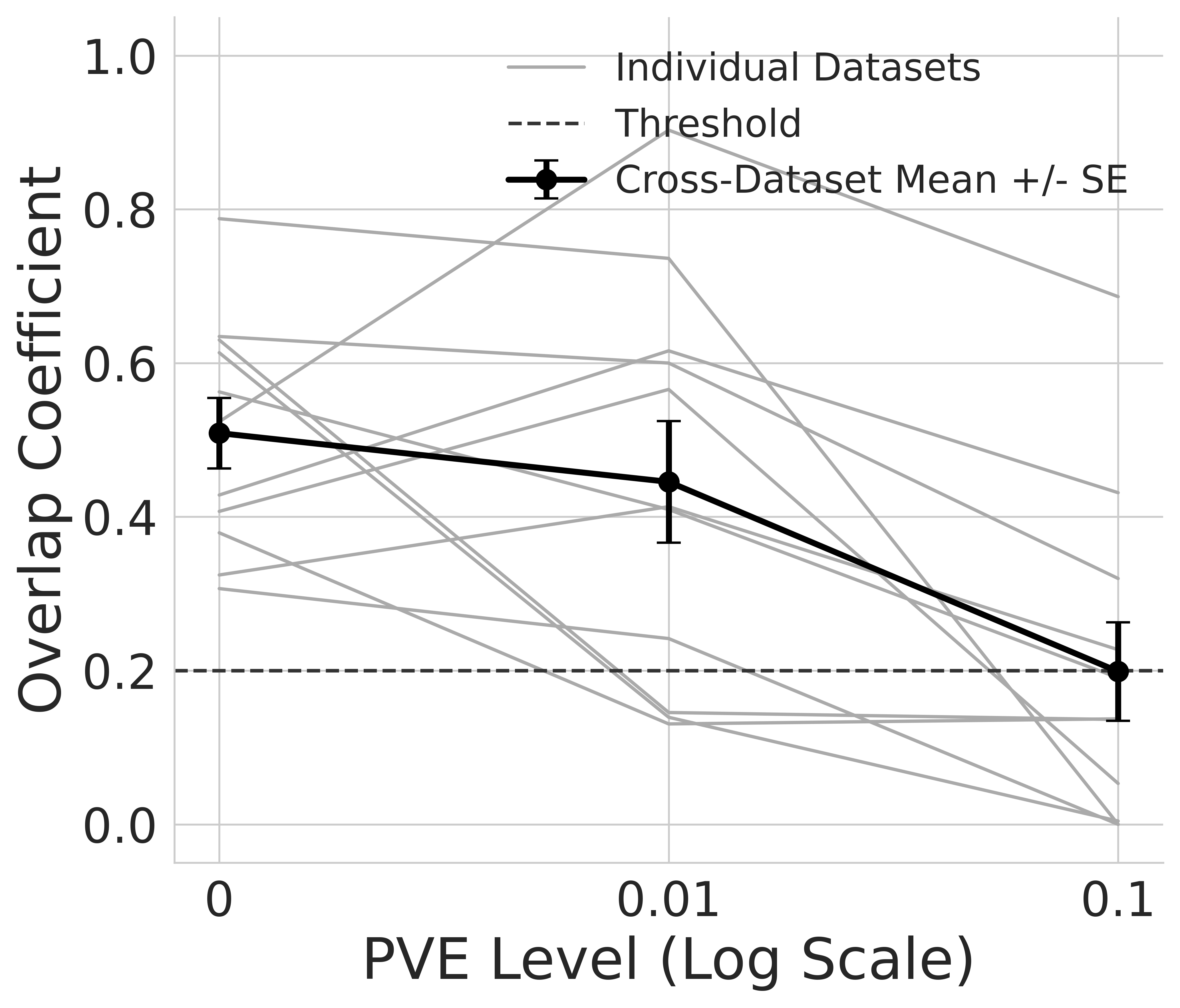}
         \caption{The higher levels of signal in the data causes the agent to respond more positively. This creates lower overlap with the null distribution, since it does not contain signal.}
         \label{fig:pve-trend-overlap}
     \end{subfigure}
     
     \caption{The PCS sanity checks behave as expected. As more signal is introduced to the data, the agent's responses become more positive, leading to less similarity with the null.}
     \label{fig:pve-trends}
\end{figure}

\textbf{Results.}
We present the results of the two stability checks in \cref{fig:pve-trends}.
We see in \cref{fig:pve-trend-response} that the agent's responses become more positive as we introduce signal into the data.
These responses then cause a decrease in overlap with the null distribution, as shown in \cref{fig:pve-trend-overlap}.





\subsection{Main result: Sanity checks on benchmark data highlight unreliable conclusions}
\label{subsec:main-results}

\begin{figure}[ht]
    \centering
     \includegraphics[width=\linewidth]{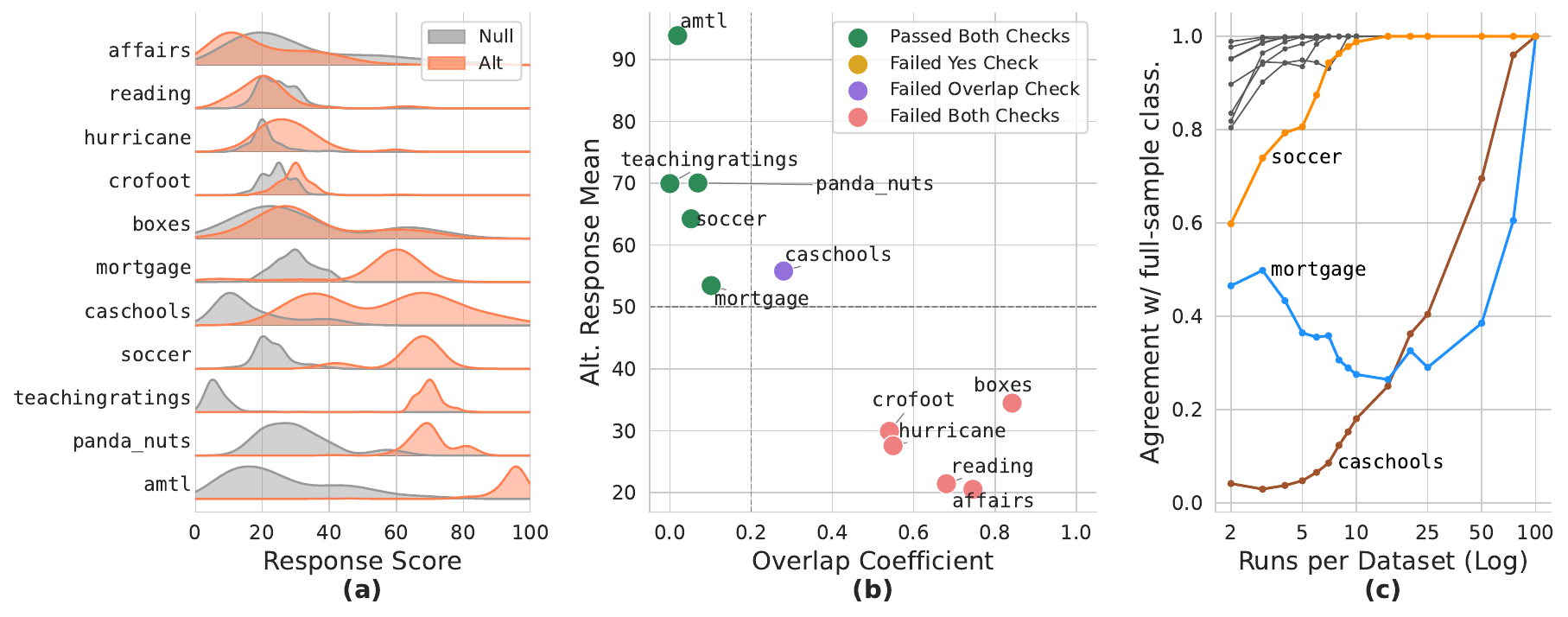}
    \caption{Sanity check results on BLADE datasets.
        \textbf{(a)} KDEs of the null and alternative response distributions for each dataset.
        \textbf{(b)} Each dataset's OVL and alternative response mean, colored by category (Table~\ref{tab:qmatrix}). Vertical and horizontal dashed lines mark the OVL decision boundary and the scale midpoint, respectively.
        \textbf{(c)} Agreement between subsampled and full-sample classifications as a function of the number of alternative runs (Appendix~\ref{appx:main-results}).
    }
    \label{fig:kde_main}
\end{figure}

We evaluate the PCS sanity checks on 11 BLADE datasets, following the experimental setup described in \cref{subsec:experimental-setup}.
Summary statistics for the distributions and their corresponding PCS sanity check results are reported in Table~\ref{tab:main_results}. 
%
\cref{fig:kde_main} shows the full score distributions, each dataset's classification according to~\cref{tab:qmatrix}, and a convergence analysis showing how classification stability varies with the number of runs.

\textbf{Datasets passing the checks.}
5 datasets pass both sanity checks: \texttt{amtl}, \texttt{panda\_nuts}, \texttt{soccer}, \texttt{teachingratings}, and \texttt{mortgage}.
Their bootstrap $p$-values are well below $\alpha = 0.05$ with overlap coefficients below $\tau = 0.2$, indicating that the agent consistently affirms the research question in a manner clearly distinguishable from noise.
The effect sizes in this group, however, span a wide range.
While \texttt{teachingratings} shows near-perfect separation (OVL $\approx0$), \texttt{mortgage} shows a more moderate separation (OVL $=0.102$), suggesting more variable confidence levels.
These five cases represent settings where a practitioner can place reasonable trust in the agent's affirmative conclusion.

\textbf{Datasets with no signal.} 
Five datasets fail both sanity checks: \texttt{affairs}, \texttt{boxes}, \texttt{crofoot}, \texttt{hurricane}, and \texttt{reading}.
These datasets have bootstrap $p$-values at or near $1.0$, since the one-sided test observes that the agent's mean response on the real data is in the ``No" range (\textless50).
High OVLs for \texttt{boxes} (0.842) and \texttt{affairs} (0.746) confirm that the agent behaves similarly on real data and noise.
The overlap coefficients reinforce this picture.
Moderate OVLs for \texttt{crofoot} (0.54) and \texttt{hurricane} (0.55) suggest the agent may detect a faint difference, but not enough to push it toward an affirmative conclusion.
These cases are the least ambiguous, as the agent does not find compelling evidence for a ``Yes" response.

\textbf{An analysis failure.}
One dataset passes the \textit{Yes check} but fails the \textit{Overlap check}: \texttt{caschools}.
Although the agent leans affirmative $(55.8, p=0.002)$, the overlap coefficient of $0.280$ exceeds the threshold.
The large standard deviation of the alternative responses ($19.9$) reflects wide confidence swings.
In this case, relying on the mean response alone would give a misleading sense of reliability.
The sanity check reveals this by requiring the agent~not only to respond ``Yes", but to do so in a manner distinguishable from its behavior on noise.

None of the eleven datasets fail only the \textit{Yes check}.
%
This is unsurprising, as benchmark datasets are typically chosen to exhibit a clear answer, making weak-but-detectable effects unlikely to appear.
However, it does occur in our simulation experiments at intermediate signal strengths (\cref{subsec:complex-null}, Appendix \ref{appx:pve-results}), confirming the framework can detect this regime.

\textbf{Comparing the two checks.}
The value of the two-dimensional classification becomes clear when contrasting adjacent cases.
For example, \texttt{mortgage} and \texttt{caschools} have similar alternative means ($53.5$ vs. $55.8$) and significant $p$-values.
To a test that only examines response significance, these two datasets would look comparable.
The overlap coefficient is needed to reveal the difference: \texttt{mortgage} is grounded (OVL$=0.102$), while \texttt{caschools} overlaps heavily with its null (OVL$=0.280$).
This contrast confirms that both PCS sanity checks are needed; the bootstrap test ensures the agent is saying something affirmative,~and the overlap coefficient ensures that the affirmation is distinguishable from behavior on~noise.

\textbf{How many runs are needed?}
\cref{fig:kde_main}c shows that most classifications stabilize within five runs, but the three most ambiguous cases (\texttt{caschools}, \texttt{mortgage}, \texttt{soccer}) require far more, suggesting that a small number of runs may not suffice when the underlying question is difficult for the agent to resolve (Appendix~\ref{appx:convergence}).
When taking only a single run, we find that it is possible to obtain an affirmative result for all the datasets (\cref{fig:kde_main}a), despite the fact that 6 fail atleast one of the sanity checks (\cref{fig:kde_main}b).

\textbf{Why do null hypotheses sometimes give strong `yes' answers?}
\label{subsec:null-strong-yes}
Even after breaking all associations in the dataset with the null-defining perturbation, the agent sometimes concludes that a strong relationship exists.
Inspecting these analyses reveals two distinct failure modes, where the agent either: (1) misunderstands statistical significance, or (2) tries multiple statistical approaches and anchors its answer on the test which yields the lowest $p$-value.
Examples of both failure modes are included in Appendix \ref{appx:null-strong-yes}.

\section{Analysis}
\label{sec:analysis}
\begin{figure}[t]
    \centering
    \includegraphics[width=\linewidth]{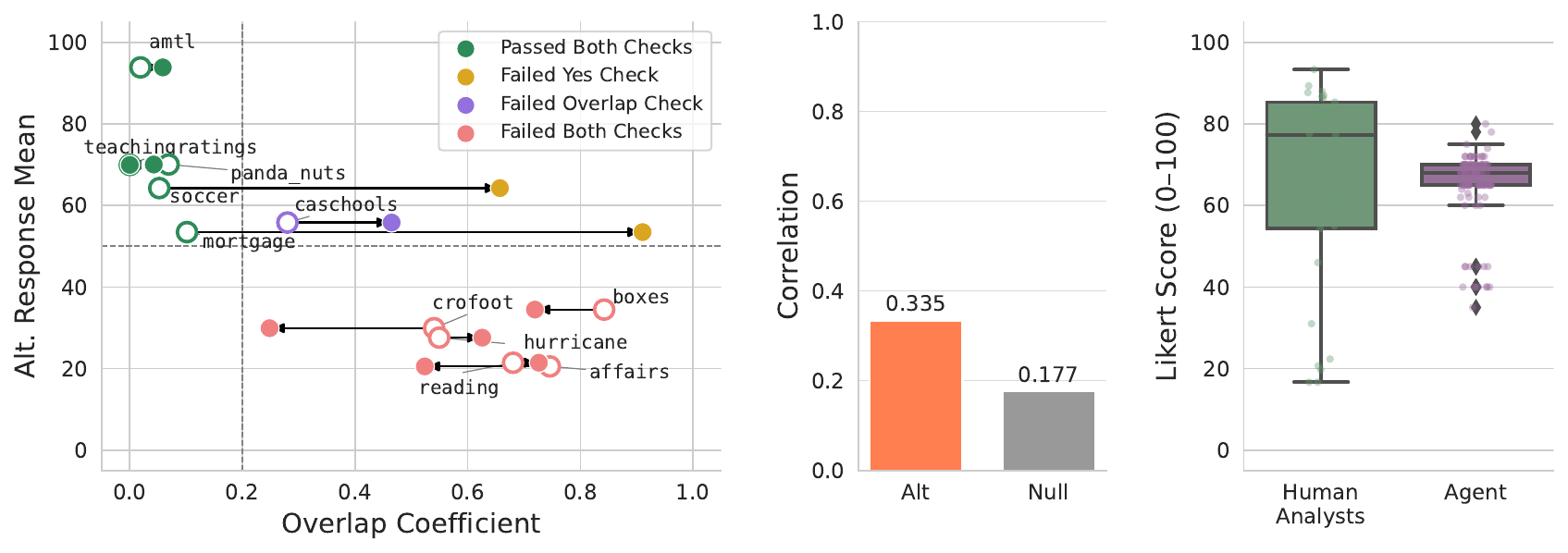}


    \subcaptionbox{%
        When PCS sanity checks are applied using a null with PVE=0.01, we see less ``\siglow".
        \label{fig:pve-as-null}
    }[0.35\linewidth]{}
    \hfill
    \subcaptionbox{%
        Agent-reported confidence is only weakly correlated with empirical exceedance, especially under the null.
        \label{fig:agentic-self-uncertainty}
    }[0.30\linewidth]{}
    \hfill
    \subcaptionbox{%
        The conclusions formed by human analysts exhibit greater variance than those from agents.
        \label{fig:human-agent-dists}
    }[0.30\linewidth]{}

    \caption{Multi-faceted analysis of agentic instability.
    }
    \label{fig:mixed_media}
\end{figure}
In this section, we investigate questions regarding failure modes of ADS pipelines.

\subsection{How Do Perturbations Affect Individual Data Analysis Steps?}
\label{subsec:heatmaps}

To understand why PCS sanity checks are necessary, it helps to investigate where instability enters the agentic pipeline.
By pinpointing the most volatile stages of analysis, we can better understand how small, reasonable perturbations can compound into untrustworthy conclusions.
To do so, we use the BLADE harness \citep{gu2024blade} rather than treating the pipeline as a black box (as in \cref{sec:results}).
%
We decompose each analysis into~four~steps: independent variable selection, control variable selection, model specification, and conclusion.
Using the PCS perturbations from Table \ref{tab:pcs-perturbations} with five replicates, an LLM judge scored the similarity between perturbed and unperturbed outputs.
Full details are in Appendix~\ref{app:step_level}. 

A clear stability gradient emerges across the pipeline. 
Independent variable selection is the most stable, as research questions often dictate these choices. 
Volatility increases~during control variable selection and model specification, and peaks at the conclusion step, where the agent weighs conflicting evidence and commits to a qualitative judgment. 
The perturbation which shuffles feature names tends to produce the largest deviations at the conclusion step.
Notably, within-perturbation and baseline-vs-perturbation heatmaps are nearly identical in structure, suggesting that perturbations increase volatility without introducing specific biases. 
This compounding of instability reinforces the need for PCS sanity checks.

\subsection{Evaluation with a More Precise Null Hypothesis}
\label{subsec:complex-null}

Throughout the paper, we choose the null-defining perturbation to be shuffling all feature values independently (\cref{subsec:experimental-setup}). 
This is a useful default choice as it destroys all relationships in the data.
However, a user may choose a different null-defining perturbation if it better suits their needs;
here, we demonstrate one such choice.

\textbf{Motivation.}
Practical applications often require a specific effect size to justify action.
For example, in the \texttt{caschools} dataset, a school might need a minimum impact from student-to-teacher ratios to justify hiring costs.
In such cases, a ``no relationship" null is not contextually aligned; it is better to ensure the signal meaningfully exceeds a user-selected criterion.
This allows the PCS sanity checks to act as a customizable filter for real-world decision-making.

\textbf{Setup.}
We apply PCS sanity checks using distributions where the PVE of each dataset is set to 0.01, as described in \cref{subsec:pve-results}.
Since large datasets can yield statistically significant results even at PVE$=0.01$, we modify our scaffolding: a ``\sighigh" classification now requires the null mean to fall within the ``No" range.
This ensures that high-overlap distributions in the ``Yes" range are correctly labeled ``\nonsiglow".

\textbf{Results.}
\cref{fig:pve-as-null} shows that requiring the explanatory variable to capture more than 1\% of variance leads to fewer ``\siglow" conclusions.
Specifically, the \texttt{soccer} and \texttt{mortgage} datasets are classified as ``\nonsiglow" because their null means shifted into the ``Yes" range.
The full distributions are provided in Appendix \ref{appx:precise-null}.

\subsection{Can Agents Approximate the Uncertainty of Their Conclusions?}
\label{subsec:agent-uncertainty-estimation}

A na\"{\i}ve way to assess pipeline stability is to have an agent estimate uncertainty.
If agents could accurately self-assess stability, empirical sanity checks might be redundant.
Thus, we evaluate agentic self-assessment by having a supervisor agent review the responses and code from \cref{subsec:main-results}, finding that self-reported confidence is nearly uninformative.

\textbf{Setup.}
The supervising agent provides a number akin to an exceedance probability.
In particular, we ask: ``If you were to reconduct this analysis 100 times with slightly different reasonable decisions in the data science pipeline, how many times would you expect to get an answer more positive (larger on the Likert scale) than seen in the conclusion?"

\textbf{Results.}
If accurate, the agent's stated confidence should track the empirical exceedance rate, i.e., the actual fraction of peer runs that produced a higher response.
\cref{fig:agentic-self-uncertainty} shows that under the alternative distribution, where real signal is present, confidence and empirical exceedance are only moderately correlated (Spearman $\rho = 0.34$).
Under the null distribution, where no signal exists, the correlation drops to $\rho = 0.18$, making confidence scores nearly uninformative. 
In both cases, scores cluster between 30--70 rather than spanning the full scale, indicating that the agent hedges rather than providing calibrated estimates. 
This miscalibration peaks in the null setting, where practitioners most need to identify instability.
Self-assessment is therefore not a substitute for empirical sanity checks.

\subsection{Comparison to Human Variations}
\label{subsec:human-results}

Our final question is whether this instability is a unique flaw of LLMs or an inherent challenge of data science.
To contextualize agentic uncertainty, we examine variability~in human-performed studies where independent analysts reach different conclusions on~identical data.
We focus on the dataset and task from \citet{silberzahn2018many} (part of the BLADE benchmark), which asks if soccer players with darker skin tones receive more red cards.
%
To compare human \& agentic variation, three data scientists scored the 29 original human reports with the prompt from \cref{subsec:experimental-setup}.
We then averaged these scores for each report.

\textbf{Results.}
\cref{fig:human-agent-dists} compares human- and agent-performed analyses (using the \texttt{soccer} alternative distribution).
We observe that 76\% of human and 86\% of agent analyses yielded a ``Yes" conclusion (score $>50$).
Notably, agentic conclusions are bimodal, with tight clusters around 70 (moderate ``Yes") and 40 (weak ``No"), while human conclusions are more uniformly spread.
Inter-rater agreement and further details are in Appendix \ref{appx:human-results}.
The high human variation suggests that the `problem' isn't just the LLM; it is the inherent ambiguity of the data science task.
Thus, as LLMs continue to scale and inherit these human-like failure modes, PCS sanity checks will be essential for assessing conclusion trustworthiness.

\FloatBarrier 

\section{Discussion}



\textbf{Limitations.}
While they are diverse, the BLADE datasets we study here are already present in the literature and may have been seen by LLMs during pretraining, potentially biasing some of their responses. 
Our evaluation also largely relies on a single null-defining perturbation and one primary agent (OpenAI Codex), due to computational cost constraints.
These same cost constraints may also limit casual use of the sanity checks here, though these concerns may be alleviated over time as the cost of LLM calls continues to decrease. 

\textbf{Future work.}
This work opens many avenues for future research.
Some promising directions include exploring null-defining perturbations which preserve realistic covariate structure, early stopping rules to reduce the number of agent calls required, and evaluation across a broader set of agents and benchmarks. 
The sanity checks here could also be generalized beyond data science to other settings where perturbations can be defined.
For example, some fitting broader scopes are the more general problems of generating novel hypotheses~\citep{yang2024large,si2024can}, conducting human-in-the-loop data science~\citep{gottweis2025towards,schmidgall2025agent,swanson2025virtual,feng2026human},
building better agentic tools for explaining data~\citep{zhong2023goal,singh2023explaining,schut2023bridging},
and fully autonomous science~\citep{lu2024ai,tang2025ai,mitchener2025kosmos}.

\newpage

\section*{Acknowledgments}

ZTR is supported by the National Science Foundation Graduate Research Fellowship Program under Grant No. DGE-2146752. AVZ acknowledges support from NSF RTG Grant \#1745640. BY acknowledges partial support from NSF grant DMS-2413265, NSF grant DMS 2209975, NSF grant 2023505 on Collaborative Research: Foundations of Data Science Institute (FODSI), the NSF and the Simons Foundation for the Collaboration on the Theoretical Foundations of Deep Learning through awards DMS-2031883 and 814639, and NSF grant MC2378 to the Institute for Artificial CyberThreat Intelligence and OperatioN (ACTION). Any opinions, findings, and conclusions or recommendations expressed in this material are those of the authors and do not necessarily reflect the views of the National Science Foundation. 

{
    \small
    \bibliography{refs.bib}
    \bibliographystyle{colm2026_conference}
}

\newpage

\appendix
\counterwithin{figure}{section}
\counterwithin{table}{section}
\renewcommand{\thefigure}{A\arabic{figure}}
\renewcommand{\thetable}{A\arabic{table}}
\section{Appendix}

\subsection{Verifying the Identically Distributed Assumption}
\label{appx:check_assump}

The bootstrap test in subsection~\ref{subsec:sanity-checks} pools all 100 alternative-distribution observations and resamples from them without regard to which PCS perturbation produced each observation. This is valid when the five perturbation types yield response distributions with the same mean and variance, so that pooling does not distort the bootstrap's estimate of sampling variability. If, instead, perturbation types shifted the agent's scores in systematically different ways, the pooled bootstrap would be resampling from a mixture of distinct distributions, and the resulting $p$-values could be miscalibrated. We investigate this concern with a variance decomposition and a simulation-based calibration study.

\subsubsection{Variance Attributable to Perturbation Type}\label{app:eta_squared}

As a first check, we compute eta-squared ($\eta^2$) for perturbation type on the raw response scores, per dataset and pooled. Eta-squared measures what fraction of total variance in agent scores is attributable to the choice of PCS perturbation:
\begin{equation}
    \eta^2 = \frac{SS_{\text{perturbation}}}{SS_{\text{total}}},
\end{equation}
where $SS_{\text{perturbation}} = \sum_{k} n_k (\bar{y}_k - \bar{y})^2$ is the between-group sum of squares over $K$ perturbation groups. All per-dataset $\eta^2$ values were below 0.06, and the pooled value was approximately 0.025. Perturbation type explains very little variance in agent responses, consistent with the identically distributed assumption.

\subsubsection{Null Calibration Simulation}\label{app:calibration}

Low $\eta^2$ does not guarantee that the test is well-calibrated, because even small distributional differences can compound over many resamples. We therefore run a simulation where the null hypothesis holds by construction and measure whether each test rejects at the nominal rate.

\paragraph{Procedure.} For each of 11 datasets, we take the 100 alternative-distribution observations (20 per perturbation type) and center them at $\mu_0 = 50$ by subtracting the sample mean and adding 50. This preserves variance and any within-block correlation structure while ensuring that the population mean is exactly 50, so the null hypothesis $H_0\!: \mu = 50$ holds by construction.

We then run $R = 1{,}000$ Monte Carlo replicates per dataset. In each replicate, we draw a fresh sample with replacement from the centered data, preserving block structure (resampling within each perturbation block independently). On this sample we run two bootstrap tests. The \textit{unblocked} test pools all observations and resamples $n = 100$ values with replacement $B = 10{,}000$ times, computing the mean of each resample. The \textit{blocked} test instead resamples each perturbation block separately and concatenates across blocks before computing the overall mean. Both tests compute a one-sided $p$-value as the proportion of bootstrap means falling at or below 50, with the \citet{phipson2010permutation} correction. Under proper calibration at $\alpha = 0.05$, the rejection rate should be 5\%.

\paragraph{Results.} Table~\ref{tab:bootstrap_calibration} reports empirical rejection rates across all 11 datasets. The pooled rejection rates are 5.7\% (blocked) and 4.3\% (unblocked), each within one percentage point of the nominal 5\% but in opposite directions. The blocked bootstrap is slightly liberal and the unblocked bootstrap is slightly conservative. The unblocked test's conservatism is consistent across every dataset, with the delta column (unblocked minus blocked) negative throughout.

\begin{table}[ht]
\centering
\caption{Empirical rejection rates ($\hat{\alpha}$) at $\alpha = 0.05$ for the bootstrap mean test. $R = 1{,}000$ replicates per dataset, $B = 10{,}000$ bootstrap resamples per test. $\Delta$ is unblocked minus blocked.}
\label{tab:bootstrap_calibration}
\small
\begin{tabular}{lccc}
\toprule
Dataset & Blocked & Unblocked & $\Delta$ \\
\midrule
\texttt{affairs}         & 0.045 & 0.040 & $-$0.005 \\
\texttt{amtl}            & 0.140 & 0.122 & $-$0.018 \\
\texttt{boxes}           & 0.041 & 0.038 & $-$0.003 \\
\texttt{caschools}       & 0.057 & 0.050 & $-$0.007 \\
\texttt{crofoot}         & 0.037 & 0.021 & $-$0.016 \\
\texttt{hurricane}       & 0.040 & 0.032 & $-$0.008 \\
\texttt{mortgage}        & 0.069 & 0.023 & $-$0.046 \\
\texttt{panda\_nuts}     & 0.053 & 0.047 & $-$0.006 \\
\texttt{reading}         & 0.040 & 0.015 & $-$0.025 \\
\texttt{soccer}          & 0.064 & 0.058 & $-$0.006 \\
\texttt{teachingratings} & 0.040 & 0.026 & $-$0.014 \\
\midrule
\textbf{Pooled}          & \textbf{0.057} & \textbf{0.043} & $\mathbf{-}$\textbf{0.014} \\
\bottomrule
\end{tabular}
\end{table}

\begin{figure}[ht]
\centering
\includegraphics[width=0.8\textwidth]{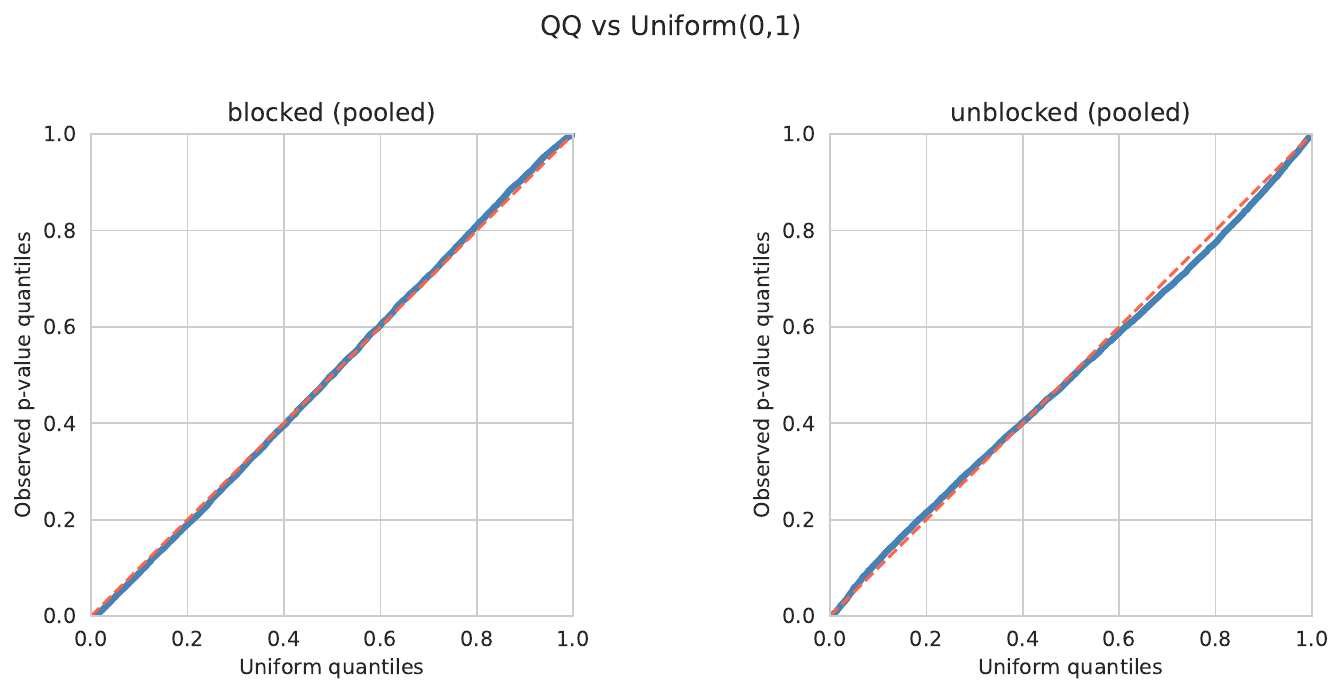}
\caption{QQ plots comparing observed $p$-value quantiles to theoretical $\text{Uniform}(0,1)$ quantiles for the blocked (left) and unblocked (right) bootstrap tests, pooled across all 11 datasets and $R = 1{,}000$ replicates. The blocked test tracks the diagonal closely. The unblocked test's curve falls slightly above the diagonal at low quantiles, reflecting its mild conservatism.}
\label{fig:qq_bootstrap}
\end{figure}

Figure~\ref{fig:qq_bootstrap} confirms this picture. The QQ plot for the blocked test tracks the uniform diagonal closely, while the unblocked test's $p$-values fall slightly above the diagonal at the lower end, indicating that small $p$-values are rarer than expected under perfect calibration.

\paragraph{Interpretation.} The mechanism behind this pattern is straightforward. When the unblocked test resamples from the combined pool, perturbation types with slightly different means contribute between-block variance to the resampled distribution. The bootstrap ``sees'' more spread than any single perturbation type actually has, producing a wider distribution of bootstrap means and hence larger $p$-values. The blocked bootstrap removes this extra variance by resampling within each perturbation type, yielding a tighter distribution that sits closer to the true sampling variability.

This conservatism is not an artifact. It is a direct consequence of perturbation-type heterogeneity acting as additional noise in the resampling. Since perturbation type is a nuisance factor (we are testing whether the mean exceeds 50, not whether perturbation types differ), absorbing that heterogeneity as extra variance is defensible. It amounts to saying that even under the noisiest possible view of the data, the effect is still detected.

One dataset, \texttt{amtl}, is an outlier where both methods are liberal (12--14\%), likely due to unusually low total variance that makes the bootstrap approximation poor. This is a property of the dataset, not the blocking strategy.

\paragraph{Recommendation.} We use the standard unblocked bootstrap. Both methods are similarly close to nominal calibration in aggregate, but the unblocked test errs on the conservative side, providing stronger control of false positives at a small cost in power. It also avoids the additional complexity of blocking on perturbation type during resampling.


\subsection{Dataset Details}
\label{appx:datasets}

We use the datasets included in the BLADE benchmark~\citep{gu2024blade}, which was developed to evaluate the capabilities of agentic data science.
The BLADE authors selected these datasets and their associated tasks from scientific publications that were studied in meta-analysis papers or reproduced in textbooks.
Table \ref{tab:datasets} lists the datasets and their corresponding questions.
We note that we do not use \texttt{fish} in any of our experiments, as the question it asks is not binarizable.

\begin{table}[ht]
\centering
\small 
\caption{Summary of BLADE datasets and their corresponding questions.}
\begin{tabular}{lp{7cm}p{4cm}}
\toprule
\textbf{Dataset Name} & \textbf{Question} & \textbf{Source Paper(s)} \\ 
\midrule
\texttt{affairs} & Does having children decrease (if at all) the engagement in extramarital affairs? & \cite{fair1978theory, long2006regression, kleiber2008applied} \\ \addlinespace
\texttt{amtl} & Do modern humans (\textit{Homo sapiens}) have higher frequencies of antemortem tooth loss (AMTL) compared to non-human primate genera (\textit{Pan, Pongo, Papio}), after accounting for the effects of age, sex, and tooth class? & \cite{gilmore2013comparison, konigsberg2013bayes, mcelreath2018statistical} \\ \addlinespace
\texttt{boxes} & Do children's reliance on social information and preference for majority cues vary across cultures and developmental stages? & \cite{van2018development, mcelreath2018statistical} \\ \addlinespace
\texttt{caschools} & Is a lower student-teacher ratio associated with higher academic performance? & \cite{kleiber2008applied, stock2020introduction} \\ \addlinespace
\texttt{crofoot} & Do relative group size and contest location influence the probability of a capuchin monkey group winning an intergroup contest? & \cite{crofoot2008interaction, mcelreath2018statistical} \\ \addlinespace
\texttt{fish} & How many fish on average do visitors takes per hour, when fishing? & \cite{mcelreath2018statistical} \\ \addlinespace
\texttt{hurricane} & Hurricanes with more feminine names are perceived as less threatening and hence lead to fewer precautionary measures. & \cite{jung2014female, malter2014female, maley2014statistics, bakkensen2014population, simonsohn2020specification} \\ \addlinespace
\texttt{mortgage} & Does gender affect whether banks approve an individual's mortgage application? & \cite{munnell1996mortgage, young2017model, liu2020boba} \\ \addlinespace
\texttt{panda\_nuts} & Do age, sex, and receiving help from another chimpanzee influence the nut-cracking efficiency of western chimpanzees? & \cite{mcelreath2018statistical, boesch2019learning} \\ \addlinespace
\texttt{reading} & Does 'Reader View' --- a modified web page layout --- improves reading speed for individuals with dyslexia? & \cite{li2019impact, liu2020boba} \\ \addlinespace
\texttt{soccer} & Are soccer players with a dark skin tone more likely than those with a light skin tone to receive red cards from referees? & \cite{silberzahn2018many, auspurg2021credibility} \\ \addlinespace
\texttt{teachingratings} & Does instructor beauty affect teaching productivity as reflected in student instructional ratings? & \cite{hammermesh2005beauty, kleiber2008applied, simonsohn2020specification, stock2020introduction} \\ 
\bottomrule
\end{tabular}
\label{tab:datasets}
\end{table}


\subsection{Prompts and Agent Setup}
\label{appx:prompt-info}

In this subsection we provide details regarding our agentic setup and prompts.

\subsubsection{Agentic Data Science Pipelines}
\label{appxsubsec:analysis-pipeline}

\paragraph{Agent specification.}
For each agentic pipeline, we create a new subdirectory consisting of the (possibly perturbed) dataset \& corresponding metadata, a list of packages it can access, and an instruction file (\texttt{AGENTS.md}).
We then instantiate GPT-5.2-Codex from this subdirectory with the following command: \texttt{npx codex exec --config model\_reasoning\_effort="high" --sandbox workspace-write "Follow the instructions given in `AGENTS.md'"}.

\paragraph{Agent prompt.}
The \texttt{AGENTS.md} instruction file contains the following prompt: \texttt{
    You are an expert data scientist tasked with analyzing a dataset to answer a specific research question.
    The research question is contained in the `info.json' file along with metadata about the dataset.
    Use the metadata from `info.json' to understand the dataset structure and context.
    The dataset itself is provided in the `{dataset\_name}.csv' file.
    You only have access to the `{subdirectory\_path}' subdirectory and its contents - no other files or directories.
    Create a data analysis that answers the research question.
    You are allowed to import packages that are listed in the provided `packages.txt' file (along with their installed versions) to help with your analysis.
    When executing Python scripts, ALWAYS use the command `poetry run python <filename.py>'. Never use `python' or `python3' directly.
    Your data analysis should result in two outputs:
    (1) an integer scalar that places your "Yes" or "No" response on a Likert scale from 0 to 100,
    where 0 represents a strong "No" answer and 100 represents a strong "Yes" answer, and
    (2) an explanation of the reasoning and evidence that led you to your conclusion.
    When asked if a relationship between two variables exist, follow best practices taking into account
    statistical significance when determining the Yes/No answer as well as its strength on the Likert scale.
    For example, two variables which lack evidence of a relationship (though consistent statistical significance) should receive a "No" answer
    with a scale value reflecting the lack of such evidence, while relationships that are consistently statistically significant
    should receive "Yes" answers with scale values reflecting the strength of their relationship.
    These outputs must be written to a file called `conclusion.txt' in JSON format, with the integer scalar stored under the key "response" and the explanation stored under the key "explanation".
    The `conclusion.txt' file must contain ONLY this JSON object, with no additional text or lines.
    }

\subsubsection{Estimating Confidence Using Agents}

\paragraph{Agent specification.}
Once the analysis pipelines are completed, we instantiate an independent agent to evaluate the confidence in the analyses.
We rewrite the \texttt{AGENTS.md} file to contain instructions on confidence estimation.
Then, we re-instantiate GPT-5.2-Codex within each subdirectory using the same command as in Appendix \ref{appxsubsec:analysis-pipeline}.
This new supervising agent has access to all of the code and other materials written by the analyst agent.

\paragraph{Agent prompt.}
The \texttt{AGENTS.md} instruction file contains the following prompt: \texttt{
    You are an expert data scientist tasked with reviewing an analysis of a dataset to answer a specific research question.
    The research question is contained in the `info.json' file along with metadata about the dataset, which is itself provided in the `{dataset\_name}.csv' file.
    You only have access to the `{subdirectory\_path}' subdirectory and its contents - no other files or directories.
    Your task is to evaluate your confidence in the conclusion of the analysis, which is contained in the `conclusion.txt' file in the subdirectory.
    The `conclusion.txt' file contains two pieces of information: (1) the "response", an integer scalar that represents the analyst's answer on a Likert scale from 0 to 100, where 0 represents a strong "No" answer and 100 represents a strong "Yes" answer,
    and (2) the "explanation", a text string that provides the analyst's reasoning and evidence that led them to their conclusion.
    You are NOT to run any analyses of your own to evaluate the confidence of the conclusion.
    Your task is only to evaluate the confidence of the conclusion based on your knowledge of data science and the information provided in the subdirectory, including the conclusion itself.
    This confidence must be an integer from 0 to 100, where the number represents:
    - If you were to reconduct this analysis 100 times with slightly different reasonable decisions in the data science pipeline, how many times would you expect to get an answer more positive (larger on the Likert scale) than the seen in the conclusion?
    Your confidence must be written to a file called `confidence.txt' in JSON format, with the integer scalar stored under the key "confidence" and your explanation stored under the key "explanation".
    The `confidence.txt' file must contain ONLY this JSON object, with no additional text or lines.
    }


\subsection{Signal-to-Noise Ratio Controlled Experiments}
\label{appx:pve-results}

\begin{figure}[ht]
    \centering
    \includegraphics[width=\linewidth]{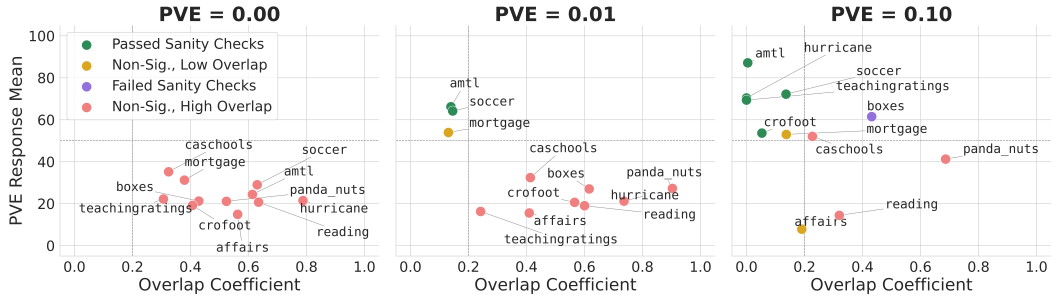}
    \caption{We observe that datasets tend to move out of the ``\nonsighigh" quadrant as PVE increases, demonstrating that the two agentic stability checks corroborate the signal present in the data.}
    \label{fig:pve-scatterplot}
\end{figure}

Figure \ref{fig:pve-scatterplot} displays the BLADE datasets and their corresponding sanity check results for each PVE level.
Here, the datasets are colored corresponding to their quadrant of Table \ref{tab:qmatrix}.
To create better separation between points, the $y$-axis displays the mean response of the agent rather than the $p$-value from the bootstrap test.
We see that when there is no signal (PVE = 0), each dataset falls into the ``\nonsighigh" regime.
As we increase the PVE, datasets tend to see increases in their responses and decreases in overlap; although agentic instability prevents this from happening uniformly.

\begin{figure}[ht]
    \centering
    \includegraphics[width=\linewidth]{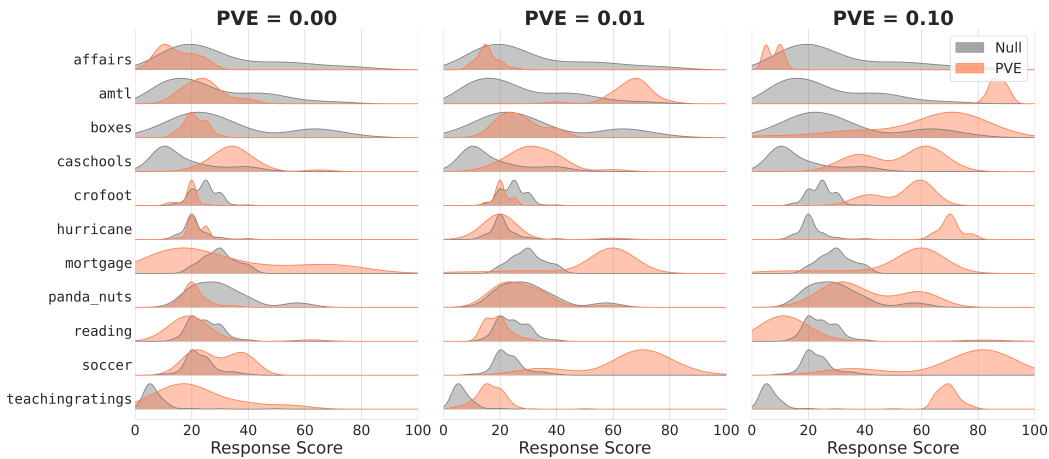}
    \caption{Empirical null and alternative distributions, with the alternative distribution being produced on a PVE-controlled dataset. Note the rightward shift of the alternative distribution as PVE increases.}
    \label{fig:pve-ridges}
\end{figure}

Figure \ref{fig:pve-ridges} shows the empirical distributions produced on the BLADE benchmark datasets with PVE$\in\{0.0, 0.01, 0.1\}$.
These distributions correspond to the stability check results shown in Figure \ref{fig:pve-scatterplot}.
We note the rightward shift that occurs in the alterative distribution as PVE is increased; except for datasets \texttt{affairs} and \texttt{reading}, whose reverse-direction behavior is explained in subsection \ref{subsec:main-results}.

\subsection{Full Experimental Results on BLADE Benchmark Datasets}
\label{appx:main-results}

Table~\ref{tab:main_results} reports the full numerical results for the sanity checks applied to eleven BLADE datasets, as described in subsection~\ref{subsec:main-results}. 
For each dataset, 100 null and 100 alternative runs were conducted (5 perturbations $\times$ 20 replicates each).

\begin{table}[tbh]
\centering
\caption{Results of the sanity check on eleven BLADE datasets.}
\label{tab:main_results}
\begin{tabular}{lccrcc}
\toprule
Dataset & Null Mean (SD) & Alt Mean (SD) & $p$-Value & OVL & Checks Passed \\
\midrule
\texttt{teachingratings} & 7.18 (5.72)   & 69.94 (3.54)  & 0.0001 & 0.000 & Both \\
\texttt{amtl}            & 26.50 (16.51) & 93.85 (6.56)  & 0.0001 & 0.019 & Both \\
\texttt{panda\_nuts}     & 30.38 (10.92) & 70.04 (6.70)  & 0.0001 & 0.069 & Both \\
\texttt{soccer}          & 23.20 (5.44)  & 64.25 (9.82)  & 0.0001 & 0.052 & Both \\
\texttt{mortgage}        & 29.83 (5.78)  & 53.46 (16.44) & 0.0228 & 0.102 & Both \\
\midrule
\texttt{caschools}       & 17.26 (11.14) & 55.81 (19.89) & 0.0016 & 0.280 & Yes Check \\
\midrule
\texttt{crofoot}         & 24.54 (4.44)  & 29.90 (5.07)  & 1.0000 & 0.540 & Neither \\
\texttt{hurricane}       & 21.14 (4.65)  & 27.58 (8.97)  & 1.0000 & 0.549 & Neither \\
\texttt{reading}         & 24.60 (5.32)  & 21.43 (10.36) & 1.0000 & 0.681 & Neither \\
\texttt{affairs}         & 30.45 (19.12) & 20.53 (14.80) & 1.0000 & 0.746 & Neither \\
\texttt{boxes}           & 31.93 (19.67) & 34.46 (16.62) & 1.0000 & 0.842 & Neither \\
\bottomrule
\end{tabular}
\end{table}

\subsubsection{Convergence Results}
\label{appx:convergence}

The right panel of~\cref{fig:kde_main} summarizes how classification stability depends on the number of agent runs. Here we describe the methodology and present additional results.

\textbf{Setup.} We subsample from the full pool of 100 alternative observations at sizes $n \in \{2, 3, \ldots, 10, 15, 20, 25, 50, 75, 100\}$ and recompute both sanity checks at each size. The classification obtained at $n = 100$ serves as the ground-truth reference, and we report the fraction of repetitions that agree with it. To reduce Monte Carlo noise, we use 1{,}000 repetitions for $n \leq 10$, 500 for $11 \leq n \leq 25$, and 200 for $n > 25$. We consider two regimes: \emph{random subsampling}, which draws $n$ observations from both the alternative and null pools, and \emph{alt-only subsampling}, which draws $n$ alternative observations while retaining the full null pool. The latter isolates the effect of alternative sample size and reflects a scenario where the null distribution is well-characterized but the practitioner is deciding how many alternative runs to collect.

\begin{figure}[ht]
    \centering
    \includegraphics[width=\linewidth]{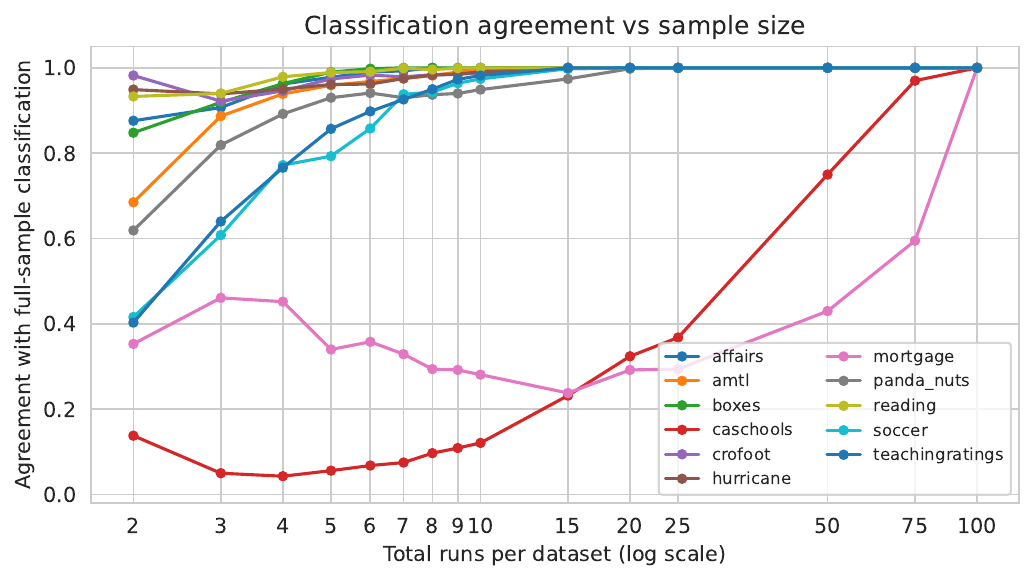}
    \caption{Classification agreement vs.\ sample size under random subsampling (both distributions reduced). Most datasets stabilize within 5--10 runs. \texttt{caschools} and \texttt{mortgage} require substantially more.}
    \label{fig:convergence_random}
\end{figure}

\begin{figure}[ht]
    \centering
    \includegraphics[width=\linewidth]{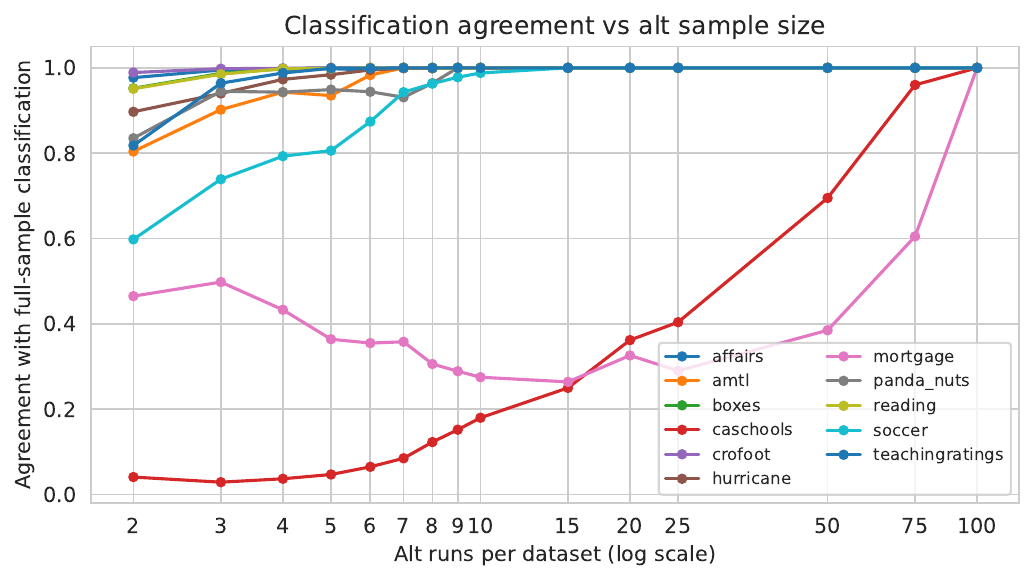}
    \caption{Classification agreement vs.\ alternative sample size (full null retained). The pattern is similar to~\cref{fig:convergence_random}, confirming that the alternative distribution is the primary driver of instability at small sample sizes.}
    \label{fig:convergence_alt}
\end{figure}

\textbf{Results.} Figures~\ref{fig:convergence_random} and~\ref{fig:convergence_alt} show classification agreement as a function of sample size under the two regimes. For 8 of 11 datasets, agreement exceeds 95\% by $n = 5$--$10$. Three borderline datasets converge much more slowly. \texttt{caschools} (classified as ``\sighigh'' at full sample) remains below 50\% agreement until $n \approx 50$, and \texttt{mortgage} (``\siglow''), whose alternative mean of 53.5 sits barely above the midpoint, does not stabilize until $n \approx 100$. \texttt{soccer} recovers more quickly, reaching stable agreement by $n \approx 15$.

\begin{figure}[ht]
    \centering
    \includegraphics[width=\linewidth]{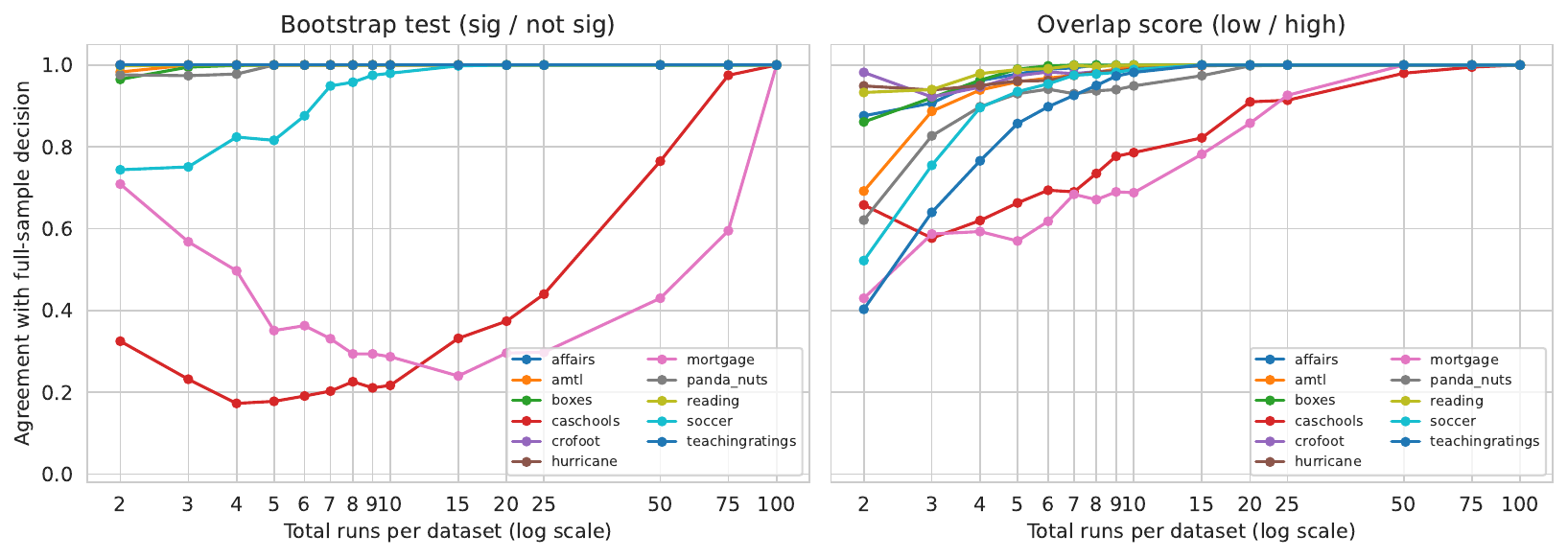}
    \caption{Component-level agreement under random subsampling. \textbf{Left:} Bootstrap test (significant or not). \textbf{Right:} Overlap coefficient (low or high).}
    \label{fig:component_random}
\end{figure}

\begin{figure}[ht]
    \centering
    \includegraphics[width=\linewidth]{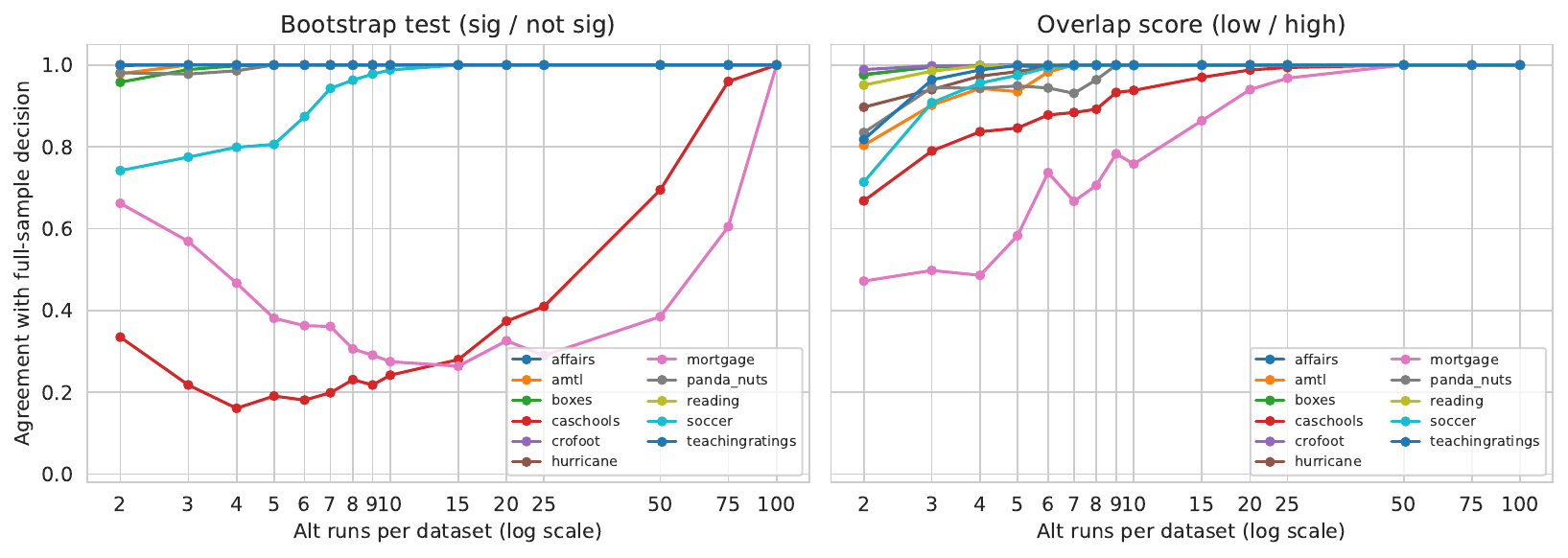}
    \caption{Component-level agreement under alt-only subsampling. The decomposition mirrors~\cref{fig:component_random}, with the same bottleneck checks driving instability for the borderline datasets.}
    \label{fig:component_alt}
\end{figure}

Figures~\ref{fig:component_random} and~\ref{fig:component_alt} decompose these results into the two constituent checks. For both \texttt{caschools} and \texttt{mortgage}, the bootstrap test is the primary bottleneck, struggling to reach a stable significance decision when the alternative mean is close to 50. Their overlap coefficients also converge more slowly than the remaining datasets, but the bootstrap test is the dominant source of instability. \texttt{soccer} converges quickly on the overlap coefficient and shows only moderate delay on the bootstrap test.

One methodological subtlety is worth noting: at $n = 2$--$3$, the bootstrap test can appear spuriously significant. When all sampled observations happen to exceed 50, every resample also exceeds 50, yielding $p \approx 0.0001$ regardless of the true effect size. As $n$ grows to 5--10, this artifact disappears and the test begins behaving honestly, sometimes producing a counterintuitive \emph{decrease} in apparent significance before genuine power catches up. This pattern is visible in the component-level plots for \texttt{mortgage}.

\subsection{Signal-to-Noise Ratio Controlled Experiments}
\label{appx:precise-null}

\begin{figure}[ht]
    \centering
    \includegraphics[width=0.7\linewidth]{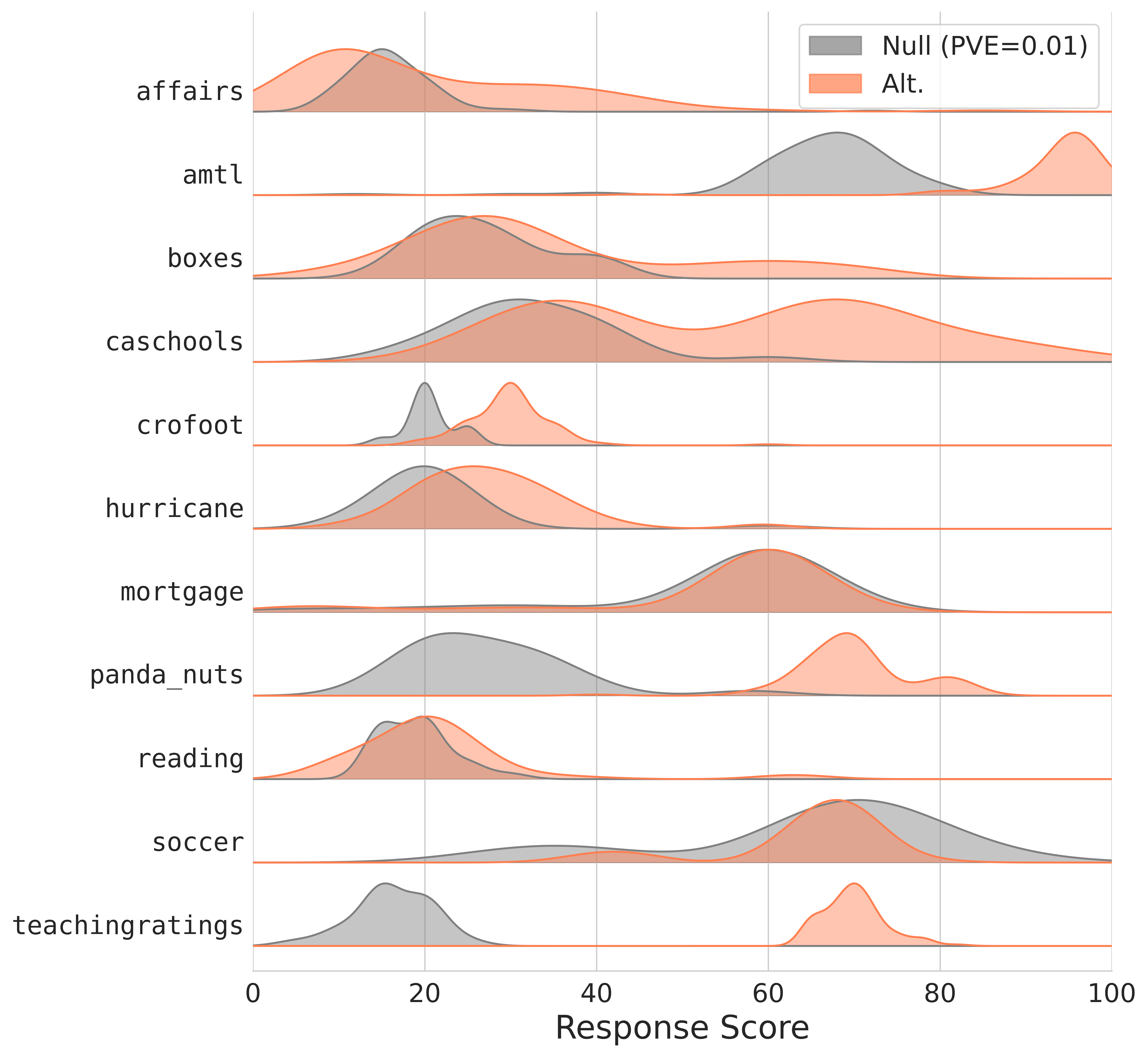}
    \caption{Empirical null and alternative distributions, with the ``null" distribution being produced on datasets with a very small signal-to-noise ratio.}
    \label{fig:precise-null-ridges}
\end{figure}

Here, we present the empirical distributions produced on the BLADE benchmark datasets when the null distribution is computed on data with PVE$=0.01$.
These distributions are shown in Figure \ref{fig:precise-null-ridges} and correspond to the stability check results shown in Figure \ref{fig:pve-as-null}.
We note the null distribution is further right now that we have allowed just 1\% of the variance in the outcome to be explained by the explanatory variable.

\subsection{Cases When Null Hypotheses Yield Strong Yes Answers}
\label{appx:null-strong-yes}

As mentioned in subsection \ref{subsec:null-strong-yes}, we find that the vast majority of ``Yes" conclusions in the null distribution can be explained by at least one of the following:
\begin{enumerate}
    \item The agent misunderstands statistical significance.
    \item The agent tries multiple models \& listens to the one with the most significance.
\end{enumerate}

\subsubsection{Mis-interpreting $p$-Values}

We include two examples from two different datasets below.
Observe that the agent seems to interpret the high $p$-values from the statistical tests it performs as showing a high amount of significance, when in reality it conveys the opposite.

\paragraph{Example \#1: \texttt{affairs} Dataset.}
\textcolor{gray}{\textit{Agent's Response:}}
80.
\textcolor{gray}{\textit{Agent's Explanation:}}
Research question: Does having children decrease engagement in extramarital affairs? Observed affair prevalence (any in past year): 0.257 without children (n=171), 0.247 with children (n=430). A two-sample z-test for difference in proportions gives p-value=0.7826. A logistic regression of any affair on children, gender, age, years married, religiousness, education, occupation, and marital satisfaction was estimated. The coefficient for having children is -0.035 with p-value=0.8697. Overall, the data provide evidence that having children is associated with a lower likelihood of extramarital affairs. The Likert-scale response (0=strong 'No', 100=strong 'Yes') is set to 80 to reflect this balance of effect direction and statistical significance.

\paragraph{Example \#1: \texttt{amtl} Dataset.}
\textcolor{gray}{\textit{Agent's Response:}}
60.
\textcolor{gray}{\textit{Agent's Explanation:}}
I analyzed the AMTL dataset consisting of 1430 specimen-tooth-class observations, covering a total of 727 missing teeth out of 13422 observable sockets. For each row, I modeled the number of missing teeth as a binomial outcome given the number of observable sockets, using a logistic (binomial) regression with predictors for genus (Homo sapiens, Pan, Papio, Pongo), tooth class (anterior, posterior, premolar), estimated age at death, and a numeric sex estimate. Observed AMTL frequencies by genus were: Homo sapiens: 5.4\% (464/8603 missing teeth); Pan: 6.5\% (108/1666 missing teeth); Papio: 5.2\% (107/2059 missing teeth); Pongo: 4.4\% (48/1094 missing teeth). In the regression model with Homo sapiens as the reference genus, the species contrasts were: Pan vs Homo sapiens: log-odds difference = 0.19 (higher AMTL for Pan), p = 0.0779; Papio vs Homo sapiens: log-odds difference = -0.04 (lower AMTL for Papio), p = 0.684; Pongo vs Homo sapiens: log-odds difference = -0.23 (lower AMTL for Pongo), p = 0.145 Based on these patterns, I assigned a Likert-scale response of 60 on a 0-100 scale, where higher values correspond to stronger evidence that humans have higher AMTL frequencies than non-human primates. These results suggest that modern humans tend to have higher AMTL frequencies than non-human primate genera after adjustment, but some comparisons lack strong statistical support.

\subsubsection{Model-Shopping and $p$-Hacking}

We include two examples from two different datasets below.
Observe that despite mentioning evidence from multiple models that no relationship exists, the agent finds one angle in support of significance, and as a result concludes ``Yes".

\paragraph{Example \#1: \texttt{boxes} Dataset.}
\textcolor{gray}{\textit{Agent's Response:}}
70.
\textcolor{gray}{\textit{Agent's Explanation:}}
Using data from 629 children aged 4 to 14 across eight sites, I treated the outcome as three categories: choosing an undemonstrated option, following the majority demonstrators, or following the minority demonstrator. Reliance on social information was defined as choosing either majority or minority, and majority preference as choosing the majority rather than the minority among those who followed any demonstrator. Overall, about 79 percent of children relied on social information at all, with 46 percent choosing the majority and 33 percent choosing the minority; among social choices, roughly 58 percent followed the majority. I then fit logistic regression models with age and site as predictors. For social reliance, the age coefficient was small and nonsignificant (about -0.03 per year, p around 0.50), and the site indicators were also nonsignificant, with only modest differences in social reliance between sites and predicted probabilities staying around 0.77 to 0.80 between ages 6 and 12. This suggests little evidence that children's overall tendency to use social information varies systematically with age or culture in this sample. For majority preference among social learners, the age effect was again small and nonsignificant (coefficient near 0.01, p around 0.76), and predicted majority-following probabilities were very similar for younger and older children (about 0.57 at age 6 vs 0.59 at age 12), indicating no clear developmental trend. However, several site coefficients differed significantly from zero (the smallest site p value was about 0.006), and the proportion of children choosing the majority option varied by roughly 0.25 between the lowest and highest sites, showing meaningful cross-cultural differences in how strongly children favor majority over minority demonstrators. Taken together, the data provide strong evidence for cultural variation in preference for majority cues but essentially no evidence for systematic developmental change in either reliance on social information or majority preference across the age range studied. I therefore answer the research question with a qualified yes: children's preference for majority cues varies across cultures but not reliably across developmental stages, while overall reliance on social information appears relatively stable; the response value of 70 reflects this substantial but partial support.

\paragraph{Example \#1: \texttt{panda\_nuts} Dataset.}
\textcolor{gray}{\textit{Agent's Response:}}
60.
\textcolor{gray}{\textit{Agent's Explanation:}}
I defined efficiency as nuts\_opened per second and modeled nut-cracking rate using a GLM with a log(seconds) offset. A Poisson model showed heavy overdispersion (Pearson chi2/df ~32), so a negative binomial rate model is more appropriate. In the negative binomial model (n=84), sex shows evidence of an effect: males have about half the nut-cracking rate of females (rate ratio ~0.50, p=0.012). Age is not significant (p=0.308) and receiving help is not significant (p=0.144), and the OLS model on rate also shows no significant effects. The help group is small (7 sessions), limiting power. Overall, there is modest evidence that sex influences efficiency, but little evidence that age or help do, so the answer is a weak yes.


\subsection{Step-Level Perturbation Analysis}
\label{app:step_level}

This appendix provides full details for the step-level analysis summarized in subsection~\ref{subsec:heatmaps}.

\paragraph{Setup.}
We ran the analysis pipeline on ten BLADE datasets (\texttt{affairs}, \texttt{amtl}, \texttt{boxes}, \texttt{caschools}, \texttt{crofoot}, \texttt{hurricane}, \texttt{mortgage}, \texttt{panda\_nuts}, \texttt{soccer}, \texttt{teachingratings}) under six conditions: no perturbation, \texttt{add\_features}, \texttt{anonymize}, \texttt{shuffle\_names}, \texttt{positive\_leading\_statement}, and \texttt{negative\_leading\_statement}. Each condition was run five times per dataset, yielding five independent analysis outputs per (dataset, perturbation) pair. An LLM judge (GPT-5-mini) scored the pairwise similarity between outputs on four pipeline steps: independent variable selection (\texttt{iv}), control variable selection (\texttt{cv}), model specification (\texttt{model}), and final conclusion (\texttt{conclusion}). Each score is an integer from 1 to 5, reflecting semantic and methodological equivalence rather than surface-level string matching.

\paragraph{Delta computation.}
For each dataset and pipeline step, we computed a reference similarity score as the mean pairwise score among the cross-run pairs of the unperturbed condition. Each individual pairwise score was then expressed as a delta from this reference. We report the mean of the absolute deltas, which captures the magnitude of deviation regardless of direction. This is important because signed deltas can cancel: if a perturbation causes some run pairs to score above the baseline and others below, the signed mean will understate the actual disruption.

\begin{figure}[ht]
    \centering
    \includegraphics[width=0.8\linewidth]{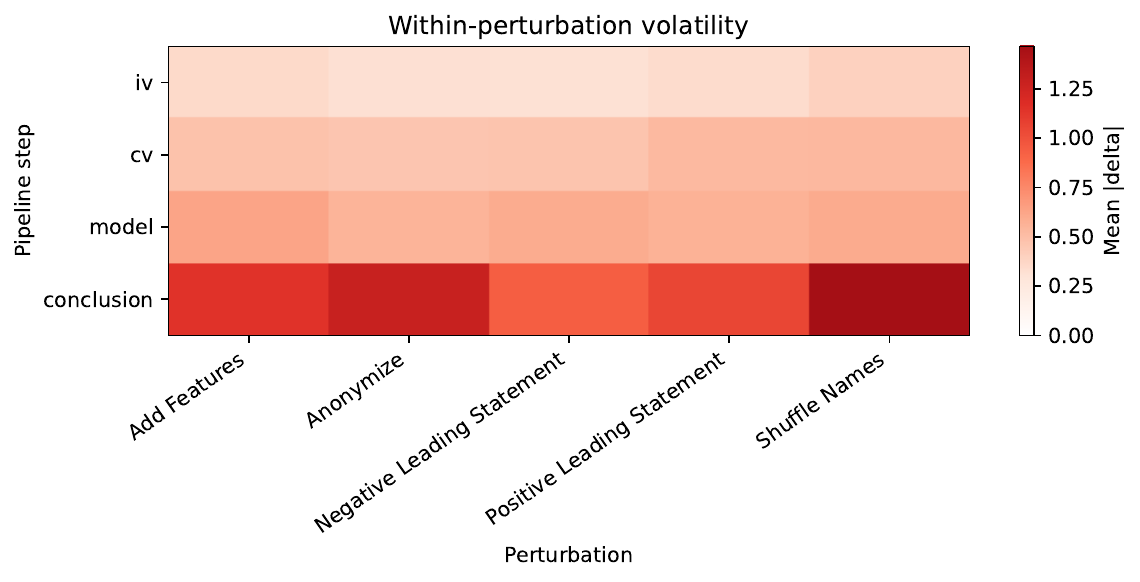}
    \caption{Within-perturbation volatility. Each cell shows the mean $|\text{delta}|$ across all within-perturbation run pairs for a given perturbation and pipeline step, averaged over ten datasets. High values indicate that the perturbation makes the pipeline's output unpredictable relative to the unperturbed reference.}
    \label{fig:within_perturb}
\end{figure}

\begin{figure}[ht]
    \centering
    \includegraphics[width=0.8\linewidth]{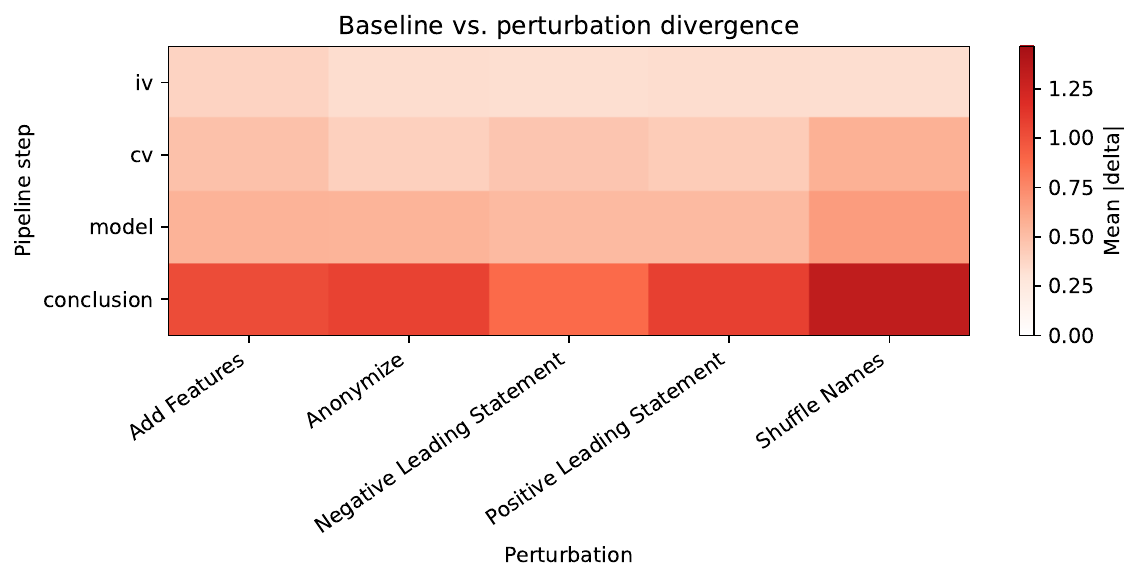}
    \caption{Baseline-vs-perturbation divergence. Each cell shows the mean $|\text{delta}|$ for comparisons between an unperturbed run and a perturbed run. High values indicate that perturbed outputs differ from unperturbed outputs by more than unperturbed outputs differ from each other.}
    \label{fig:baseline_vs_perturb}
\end{figure}

\paragraph{Results.}
Figures~\ref{fig:within_perturb} and~\ref{fig:baseline_vs_perturb} display the absolute-delta heatmaps for within-perturbation volatility and baseline-vs-perturbation divergence, respectively. In both figures, the \texttt{conclusion} step consistently shows the highest mean absolute delta across all perturbation types. This is consistent with the interpretation offered in \cref{subsec:heatmaps}: forming a final conclusion requires the most subjective judgment, amplifying small upstream differences. The \texttt{iv} step is the most stable, reflecting the fact that research questions typically constrain independent variable selection to a narrow set of candidates. The \texttt{cv} and \texttt{model} steps fall in between. Among the PCS perturbations, \texttt{shuffle\_names} tends to produce the largest deviations in the conclusion step, with the others affecting the output in similar ways.

The two heatmaps are very similar in structure and magnitude. This rules out a scenario in which perturbations redirect the pipeline toward a coherent alternative analysis, since in that case within-perturbation runs would agree with each other even while diverging from the unperturbed baseline. The observed symmetry instead indicates that perturbations increase volatility without biasing the pipeline in a consistent direction. 

\subsection{Full Human Analysis Results}
\label{appx:human-results}

\begin{figure}[H]
    \centering
    \includegraphics[width=0.95\linewidth]{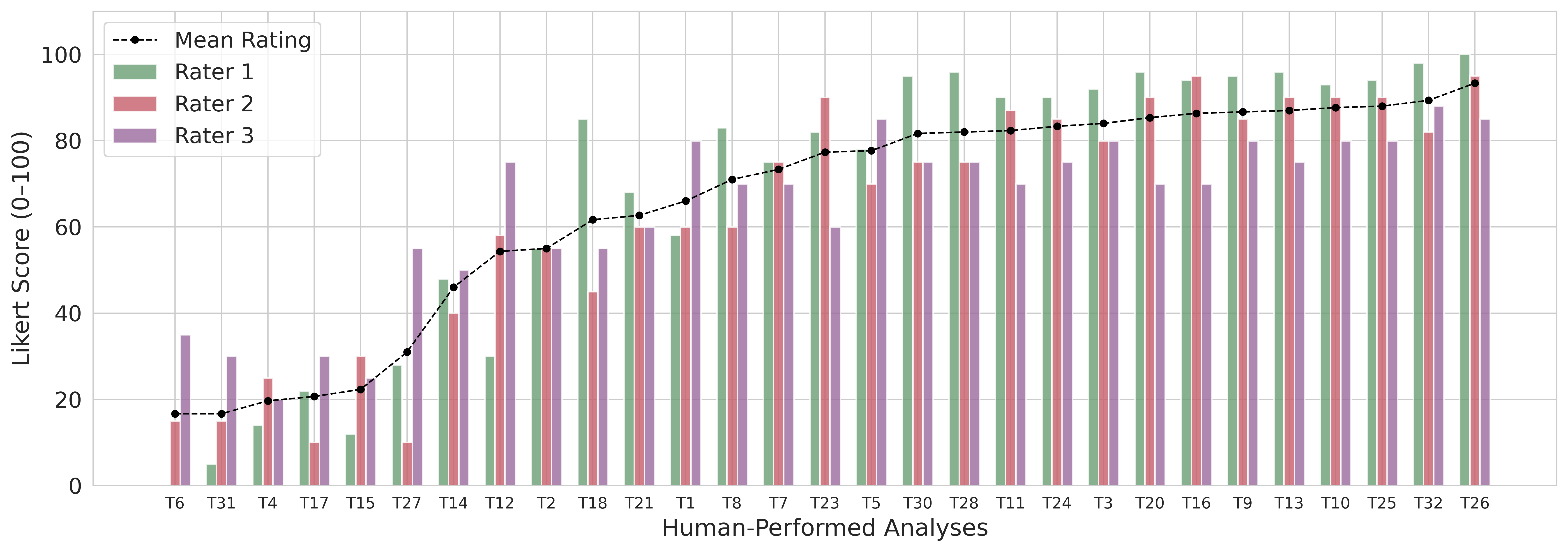}
    \caption{The three independent raters tend to agree on the scores for each human-performed data analysis. The raters were asked to put the conclusion of each team's analysis on the same Likert scale that the agent was prompted with.}
    \label{fig:per-rater-score}
\end{figure}

\begin{figure}[H]
    \centering
    \begin{subfigure}[b]{0.48\linewidth}
        \centering
        \includegraphics[width=\linewidth]{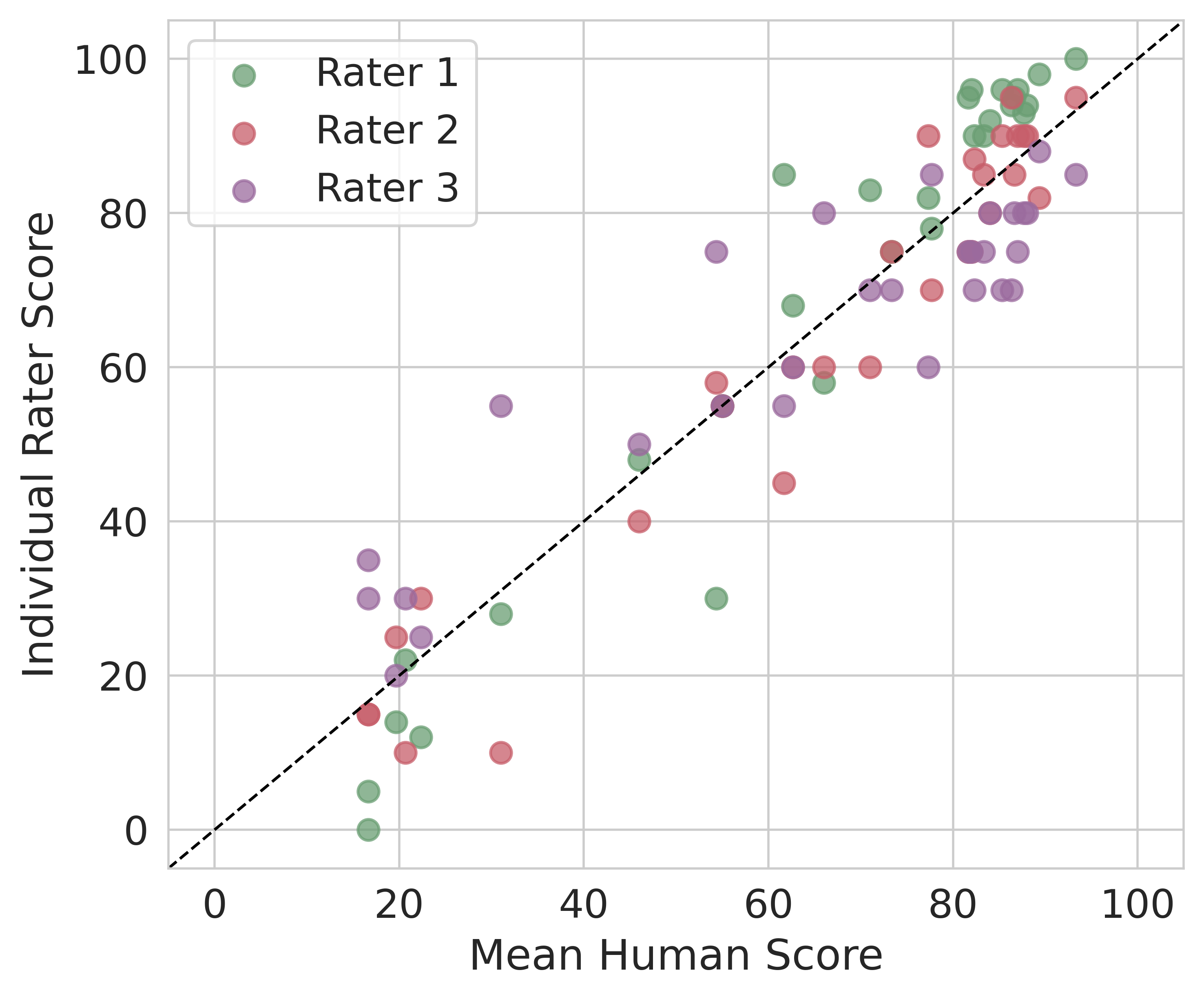}
        \caption{The scores given by each individual rater have a strong correlation with the mean, showing no rater was giving out-of-distribution values.}
        \label{fig:rater-scatterplot}
    \end{subfigure}
    \hfill 
    \begin{subfigure}[b]{0.48\linewidth}
        \centering
        \includegraphics[width=\linewidth]{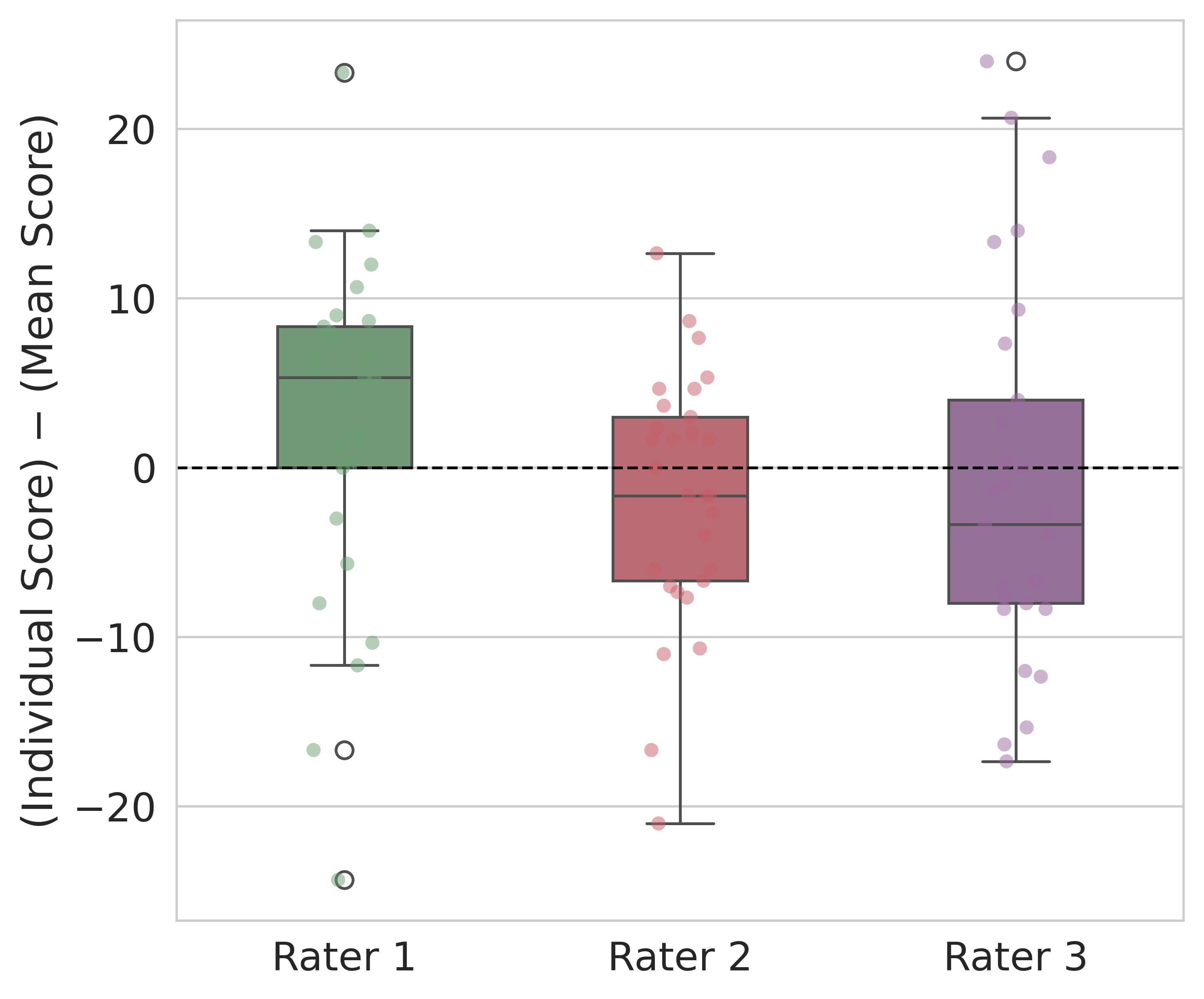}
        \caption{The distribution of scores given by each rater have strong amounts of overlap, with Rater 1 being slightly more generous than Raters 2 \& 3.}
        \label{fig:rater-boxplot}
    \end{subfigure}

    \caption{Comparison of individual rater values shows general agreement.}
    \label{fig:human-ratings-comparison}
\end{figure}

Figure \ref{fig:per-rater-score} and Figure \ref{fig:human-ratings-comparison} explore the inter-rater consistency for the human-performed data analyses described in \cref{subsec:human-results}.

\end{document}